\documentclass[article,doublecolumn]{IEEEtran}
\usepackage[latin9]{inputenc}
\usepackage{color}
\usepackage{textcomp}
\usepackage{amsmath}
\usepackage{amssymb}
\usepackage{graphicx}

\makeatletter

\providecommand{\tabularnewline}{\\}

\@ifundefined{showcaptionsetup}{}{%
 \PassOptionsToPackage{caption=false}{subfig}}
\usepackage{subfig}
\makeatother

\begin{document}

\title{On the Application of Support Vector Machines to the Prediction of
Propagation Losses at 169 MHz for Smart Metering Applications}

\author{Martino Uccellari, Francesca Facchini, Matteo Sola,~\IEEEmembership{Member,~IEEE},
Emilio Sirignano, \\
Giorgio M. Vitetta,~\IEEEmembership{Senior Member,~IEEE}, \\
Andrea Barbieri and Stefano Tondelli\thanks{Martino Uccellari, Matteo Sola, Emilio Sirignano and Giorgio M. Vitetta
are with the Department of Engineering ``Enzo Ferrari'', University
of Modena and Reggio Emilia, Modena, Italy (e-mail: matteo.sola@unimore.it,
emilio.sirignano@unimore.it, giorgio.vitetta@unimore.it). Francesca
Facchini is with the University of Modena and Reggio Emilia, where
she is enrolled in an industrial PhD program supported by the company
CPL Concordia (Concordia, Modena, Italy; e-mail: francesca.facchini@unimore.it).
Andrea Barbieri and Stefano Tondelli are with the company CPL Concordia,
that has funded this work under the contract ``Wireless channel modelling
in the 169 MHz band''.}}
\maketitle
\begin{abstract}
\textit{\textcolor{black}{\emph{Recently, the need of deploying }}}new
wireless networks for smart gas metering \textit{\textcolor{black}{\emph{has
raised the problem of radio planning in the }}}169 MHz band. Unluckily,
software tools commonly adopted for radio planning in cellular communication
systems cannot be employed to solve this problem because of the substantially
lower transmission frequencies characterizing this application. In
this manuscript a novel data-centric solution, based on the use of
support vector machine techniques for classification and regression,
is proposed. Our method requires the availability of a limited set
of received signal strength measurements and the knowledge of a three-dimensional
map of the propagation environment of interest, and generates both
an estimate of the coverage area and a prediction of the field strength
within it. Numerical results referring to different Italian villages
and cities evidence that our method is able to achieve good accuracy
at the price of an acceptable computational cost and of a limited
effort for the acquisition of measurements in the considered environments.\end{abstract}

\begin{IEEEkeywords}
Machine learning, support vector machine\textit{\textcolor{black}{\emph{,
measurement, }}}received signal strength,\textit{\textcolor{black}{\emph{
radio coverage, smart metering, open data.}}}
\end{IEEEkeywords}

\section{Introduction}

In recent years an increasing attention has been paid to the development
of advanced systems for electricity, gas and water metering. Various
\emph{advanced metering infrastructure} (AMI) solutions are currently
available on the market. All of them are based on a hierarchical topology,
in which multiple indoor \emph{smart meters} are connected to a \emph{gateway}
(also called \emph{data collector} or \emph{concentrator}), which
collects their measurements and send them to an infrastructure where
the acquired data are stored and processed. Recently, the \emph{Open
Metering System} group has proposed the adoption of the \emph{Wireless
M-Bus} technology for metering scenarios and, in particular, has recommended
its use in smart meters \cite{EN13757-4:2005}, \cite{EN13757-4:2013},
because of its low energy requirements and its use of sub-GHz transmission
bands for long range communications. Note that, whereas ref. \cite{EN13757-4:2005}
prescribed the use of the 868 MHz and 468 MHz bands, ref. \cite{EN13757-4:2013}
has added new transmission modes providing lower data rates in the
169 MHz band, which has been recently relocated by the European Union
to specific applications for smart metering \cite{ERC/REC70-03}.
Since outdoor radio-wave propagation is easier at lower frequencies,
the 169-MHz band is currently adopted for automated meter-reading
purposes in various European countries (e.g. in Italy, France, and
Spain). In particular, point to multipoint networks operating in the
169 MHz band have been proposed for the Italian gas market by the
\emph{Italian Gas Committee} (CIG). The design of new AMIs for gas
resources in that band has unavoidably raised the important problem
of radio planning at 169 MHz; in fact, this problem cannot be solved
by means of the software tools devised for cellular networks because
of the substantially higher transmission frequencies employed in such
networks. Moreover, the development of new tools requires an in-depth
understanding of the propagation losses experienced over the wireless
links connecting data concentrators and smart meters. Unluckily, few
results are available in the technical literature about such losses
in the 169 MHz band and the available prediction methods for those
frequencies may not provide a unique outcome. Recently, the problem
of assessing the propagation losses at 169 MHz has been investigated
by \cite{Fuschini} for its practical relevance. The interesting results
illustrated in that manuscript refer, however, to the \emph{outdoor-to-indoor}
propagation losses only (in particular, to the so called \emph{building
penetration losses}), that is to the main signal attenuation produced
by the obstruction of the walls of the building in which a given smart
meter is installed; unluckily, no result is provided about the \emph{prediction
of} \emph{outdoor propagation losses}, i.e. about the mean attenuation
that would be experienced over a given link at a arbitrary distance
if the involved smart meter was placed outdoor. In this manuscript
we tackle the last problem and develop a novel solution to the problem
of predicting outdoor propagation losses at 169 MHz. 

It is well known that, generally speaking, the prediction of wireless
path losses can be based on \emph{a priori models} or on \emph{models
based on a set of measurements acquired in various locations of that
scenario} \cite{Phillips_2013}. The models of the first category
allow engineers to make predictions exploiting available prior knowledge
only, without making use of explicit measurements; these include various
models based on different analytical methods (and, possibly, including
corrections factors deriving from measurements acquired in one or
more environments, like the well known \emph{Hata-Okumura model} \cite{Okumura},
\cite{Hata}) and the so called \emph{ray-tracing models} \cite{Rautiainen}.
These models are most appropriate for making predictions when it is
impossible or difficult to obtain measurements and are characterized
by different degrees of accuracy, knowledge of propagation environment
and complexity (for instance, ray-tracing models may offer accurate
predictions at the price of their large computation and data requirements).
On the contrary, the second category is based on a data-centric approach
known as \emph{modelling with measurements} \cite{Phillips_2013};
the models it includes rely on the idea that that there is no single
set of a priori constants, functions, or data able to provide an accurate
description of a new propagation environment. For this reason, the
evaluation of reliable predictions requires collecting measurements
at multiple locations and fitting them to estimate propagation losses
at different locations. Note that this approach can result in truthful
predictions at the price of a moderate effort in data acquisition
and of an acceptable computational load; consequently, it may provide
useful solutions for practical applications. Unluckily, a limited
number of papers about this category of models is available in the
technical literature (see \cite{Phillips_2013} and references therein)
and, as far as we know, none of them addresses the problem of predicting
the propagation losses in the 169 MHz band. This has motivated our
interest in developing a specific methodology for solving this problem.
The approach we adopted in our work has been inspired by some previous
work on the application of \emph{machine learning techniques} \cite{Bishop}
and, in particular, of \emph{supervised learning algorithms} to the
problem of field strength prediction (e.g., see \cite{Lagan=0000E0_2009}-\cite{Piacentini_2010})
and is based on the use of: a) a nonlinear model for representing,
respectively, the \emph{received signal strength} (RSS) measurements
collected at ground level in multiple locations of a given scenario;
b) a significant (but small) set of \emph{features} depending on the
type of scenario in which RSS is predicted and extracted from a \emph{three-dimensional}
(3D) geometrical representation of the considered propagation environment;
c) \emph{support vector machine} (SVM) algorithms to estimate the
system coverage area and predict the dependence of RSS on receiver
location inside the coverage area itself. The proposed method can
be easily implemented using public SVM libraries, requires the use
of low cost commercial hardware for data acquisition, and entails
an acceptable computational load for the considered application; moreover,
as evidenced by our numerical results, it is able to provide accurate
predictions in different propagation scenarios. For this reason, we
believe that it represents the key for the development of efficient
software tools for radio planning of AMIs operating at 169 MHz. 

It is important to mention that the application of SVM methods to
the field strength prediction is not new and that, in particular,
has been previously investigated in \cite{Lagan=0000E0_2009}, \cite{Piacentini_2010}
and \cite{Robs14}. However, our contribution represents a significant
advancement with respect to previous work since: a) the numerical
results illustrated in those manuscripts refer to substantially higher
frequencies (900 MHz, 947 MHz and 853.71 MHz, respectively); b) each
of those manuscripts refers to a single urban scenario, whereas various
propagation environments have been taken into consideration in assessing
the performance of our method; c) the set of features taken into consideration
in our work differs from that adopted in previous works and is small
(so that the use of \emph{dimensionality reduction techniques}, like
those considered in \cite{Piacentini_2010}, is not required); d)
unlike the technical solutions proposed in those manuscripts, our
approach combines SVM-based \emph{classification} and \emph{regression}~algorithms
in field strength prediction; e) unlike \cite{Lagan=0000E0_2009},
our results are entirely based on real measurements.

The remaining part of this manuscript is organized as follows. In
Section \ref{sec:SVM} the prediction of propagation losses in the
169 MHz band is formulated as a nonlinear regression problem and a
specific solution based on the use of SVM-based classification and
regression techniques is illustrated. A description of the environments
considered in our measurement campaigns and of the measurement tools
adopted in the acquisition of RSS data is provided in Section \ref{sec:Instr}.
In Section \ref{sec:Dataproc} we analyse the set of features we selected
for RSS prediction and provide various details about the implementation
of the specific software tool we developed for radio planning. Various
numerical results are illustrated and commented in Section \ref{sec:Numerical-and-experimental}.
Finally, some conclusions are offered in Section \ref{sec:Conclusions}.

\section{Support Vector Machines for the Prediction of the Propagation Losses\label{sec:SVM}}

In this Section we first formulate the problem of field strength prediction
in a specific environment. Then, we propose a specific solution based
on SVM methods for classification and regression.

\subsection{Problem statement\label{sub:Problem-statement}}

Following \cite{Fuschini}, the total propagation loss affecting the
wireless link between an indoor gas meter and a data concentrator
can be expressed (in dB) as
\begin{equation}
L_{TOT}=\left\langle PL\right\rangle +X_{SH}+X_{BPL}+\left\langle IL\right\rangle ,
\end{equation}
where $\left\langle PL\right\rangle $ is the \emph{path loss} (i.e.,
the mean attenuation that would be experienced over the given link
at the same distance if the meter was placed outdoor), whereas $X_{SH}$,
$X_{BPL}$ and $\left\langle IL\right\rangle $ represent the \emph{outdoor
shadowing}, the \emph{building penetration loss} (BPL) and the mean
\emph{installation loss} (IL; this term is due to the additional signal
shielding that originates from meter housing), respectively. Specific
indications about the modelling of $X_{BPL}$ and $\left\langle IL\right\rangle $
are provided in \cite{Fuschini}; for this reason, in our work we
focus on the problem of predicting the quantity 
\begin{equation}
L_{P}=\left\langle PL\right\rangle +X_{SH},\label{eq:Lp}
\end{equation}
that represents the so called \emph{outdoor propagation loss}. Note
that, if $L_{P}$, $X_{BPL}$ and $\left\langle IL\right\rangle $
were known, the link budget equation
\begin{equation}
P_{RX}=P_{TX}+G_{TX}+G_{RX}-L_{P}-X_{BPL}-\left\langle IL\right\rangle ,\label{eq:link_bud_1}
\end{equation}
could be used to assess the field strength at an arbitrary distance
and, consequently, to assess the \emph{system coverage area} (i.e.,
the geographical area served by the considered concentrator); here,
$P_{RX}$ ($P_{TX}$) and $G_{TX}$ ($G_{RX}$) denote the receive
(transmit) power\footnote{Received power should be always interpreted as a \emph{local mean},
since multipath fading effects are averaged out in our measurement
procedure.} (expressed in dBm) and the transmit (receive) antenna gain (expressed
in dB), respectively. However, in the following, since we are interested
in $L_{P}$ (\ref{eq:Lp}), and $P_{TX}$, $G_{TX}$ and $G_{RX}$
are perfectly known, we take into consideration the specific problem
of predicting $P_{RX}$ in an outdoor scenario. For this reason, in
this case, the link budget equation 
\begin{equation}
P_{RX}=P_{TX}+G_{TX}+G_{RX}-L_{P}\label{eq:link budget}
\end{equation}
should be employed in place of (\ref{eq:link_bud_1}). It is also
worth mentioning that: a) the sum $G_{TX}+G_{RX}$ is quite small
in the considered scenario since a low efficiency should be expected
for the antenna at the meter side; b) estimating the system coverage
area means delimiting the area that surrounds the considered concentrator
and such that, if a given meter is placed inside it, the inequality
\begin{equation}
P_{RX}\geq P_{RX-SENS}\label{eq:inequality_cov_area}
\end{equation}
holds, where \emph{$P_{RX-SENS}$} is the \emph{receiver sensitivity
threshold} (\emph{$P_{RX-SENS}=-119\,\mbox{dBm}$} in this case);
from the last expression and (\ref{eq:link budget}) it is easily
inferred that, from a mathematical viewpoint, this is equivalent to
delimiting the area in which the propagation loss $L_{P}$ satisfies
the inequality
\begin{equation}
L_{P}\leq-P_{RX-SENS}+P_{TX}+G_{TX}+G_{RX}.\label{eq:inequality_cov_area_bis}
\end{equation}
In our work the estimation of the coverage area and the prediction
of the field strength within it are based on the following data: 
\begin{enumerate}
\item A 3D geometrical representation of the considered propagation environment.
As it will become clearer later, the required representation is provided
by typical topographic maps in digital format; in fact, these usually
contain essential information about building size and position, and
about contours connecting points at the same elevation\footnote{Vector maps available in topographic databases usually contain the
required information.} (if the considered area is hilly or mountainous); as it will become
clearer later, these information are needed for the computation of
specific position-dependent \emph{features}, whose knowledge is required
by the developed SVM-based algorithms for all the locations at which
measurement have been collected and field strength is predicted. Note
also that typical maps usually provide various information about roads;
such information may be required in our approach, but not for the
evaluation of the above mentioned features. In fact, as it will be
explained in more detail in Paragraph \ref{sub:software-tool}, road
information can be exploited in some cases to improve the quality
of position estimates provided by the \emph{global positioning system}
(GPS) receiver embedded in our measurement hardware.
\item A set $\left\{ m_{i};i=1,2...,N\right\} $ collecting $N$ power measurements
acquired at $N$ distinct locations of the considered environment
(the experimental set-up adopted in our measurement campaign is described
in detail in Paragraph \ref{sub:Datacoll}). 
\end{enumerate}
In our work the nonlinear model
\begin{equation}
m_{i}=f\left(\mathbf{x}_{i}\right)+n_{i}\label{nonlinearmodel}
\end{equation}
is adopted for the $i$-th measurement (with $i=1,2,...,N$); here,
the function $f\left(\cdot\right)$ expresses the \emph{nonlinear}
dependence of the considered power on a $m$-dimensional vector $\mathbf{x}_{i}=[x_{i}^{(1)},x_{i}^{(2)},...,x_{i}^{(m)}]^{T}$
collecting $m$ distinct features\emph{}\footnote{Generally speaking, the set of features should include all the geometrical
and physical parameters that may significantly influence field strength
at an arbitrary position; the specific features we have selected for
our measurement model (\ref{nonlinearmodel}) are defined in Paragraph
\ref{sub:features}.} and $n_{i}$ is the additive noise affecting the $i$-th measurement
(noise is mainly due to the measurement equipment and to the non ideal
spatial averaging of the received power, and is assumed to be statistically
independent of the position $\mathbf{p}=[x,y,z]^{T}$ at which data
are acquired). As it will become clearer later, the selected features
depend on the coordinate vector $\mathbf{\mathbf{p}}_{i}=[x_{i},y_{i},z_{i}]^{T}$
identifying the $i$-th measurement location (in practice, an estimate
$\mathbf{\mathbf{\tilde{p}}}_{i}=[\tilde{x}_{i},\tilde{y}_{i},\tilde{z}_{i}]^{T}$,
generated by a GPS device, is available for this vector) and on various
characteristics of the propagation environment (including the transmitter
location), as illustrated in detail in Paragraph \ref{sub:features}.
It is also important to keep in mind that any meaningful power measurement
is lower bounded by the receiver sensitivity (in other words, $m_{i}\geq P_{RX-SENS}$
inside the coverage area) and that the conventional value $P_{NC}=-120\,\mbox{dBm}$
is assigned to all the measurements acquired at the positions in which
the transmitted signal is not correctly received or is not received
at all (in other words, outside the coverage area).

Given the information and the measurement models illustrated above,
we are interested in developing a procedure for solving the following
two correlated problems: a) identifying the system coverage area;
b) evaluating an approximate expression of the function $f\left(\cdot\right)$
appearing in (\ref{nonlinearmodel}), so that a field strength map
(or, equivalently, a propagation loss map) can be generated. Note
that, on the one hand, the first problem can be interpreted as a \emph{binary
classification problem}, since it concerns differentiating the cases
in which a reliable wireless link can be established from those in
which this is impossible. On the other hand, the second one can be
seen as a \emph{nonlinear regression problem} \cite{Bishop}, since
it concerns the prediction of the field strength at positions different
from those for which measurements are available. In the following
Paragraph we show that both problems can be tackled resorting to SVM
techniques and that the solution to the first problem provides some
useful information that can be exploited in the second problem.

\subsection{Application of Support Vector Machines to the Prediction of System
Coverage and Field Strength\label{sub:Theory}}

As already mentioned above, the first part of our approach concerns
the identification of the system coverage area; this problem can be
formulated as a binary classification problem. In fact, the available
information, represented by the data set $S^{(m)}\triangleq\left\{ \left(\mathbf{x}_{i},m_{i}\right),i=1,2,...,N\right\} $,
can be employed to generate the new set $S^{(z)}\triangleq\left\{ \left(\mathbf{x}_{i},z_{i}\right),i=1,2,...,N\right\} $,
where $z_{i}$ represents a binary \emph{categorical attribute} (or
\emph{class label}) specifying the absence or presence of radio coverage
at the location $\mathbf{x}_{i}$ and taking on the values $\left\{ \pm1\right\} $
only; in particular $z_{i}=+1$ ($z_{i}=-1$) identifies the presence
(lack) of radio coverage and is associated with $m_{i}\geq P_{RX-SENS}$
($m_{i}=P_{NC}$), since the received signal strength is (not) large
enough to ensure a reliable reception. In our approach, the set $S^{(z)}$
is partitioned into the \emph{training set} $S_{train}^{(z)}\triangleq\{(\mathbf{x}_{i},z_{i}),i=1,2,...,N_{train}^{(z)}\}$
and in the \emph{test set} $S_{test}^{(z)}\triangleq\{(\mathbf{x}_{i},z_{i}),i=N_{train}^{(z)}+1,N_{train}^{(z)}+2,...,N_{train}^{(z)}+N_{test}^{(z)}\}$,
consisting of $N_{test}^{(z)}$ and $N_{train}^{(z)}\triangleq N-N_{test}^{(z)}$
points, respectively. Then, the set $S_{train}^{(z)}$ is processed
by a $C$ - s\emph{upport vector classification} (C-SVC) algorithm
\cite{Cortes_1995}, \cite{Boser_1992}, that aims at identifying
an \emph{hyperplane} $H$ (in a proper $d$-dimensional Euclidean
space $\mathcal{H}$) separating the positive examples (i.e., those
associated with a categorical attribute equal to +1) from the negative
ones (i.e., those corresponding to a categorical attribute equal to
-1), while keeping the hyperplane \emph{margin} (i.e., the hyperplane
distance from the closest positive and negative examples) as large
as possible and the number of classification errors\footnote{Errors are due to the fact that, generally speaking, in the considered
problem the data collected in $S^{(y)}$ are \emph{not separable},
i.e. an hyperplane separating the points with opposite categorical
attributes does not exist. } in the training process as small as possible; note that, generally
speaking, the points of $H$ satisfy the equation
\begin{equation}
\mathbf{w}\cdot\mathbf{\phi}\left(\mathbf{x}\right)+b=0,\label{eq:hyper_eq}
\end{equation}
where $\mathbf{\phi}\left(\mathbf{x}\right)$ represents a proper
mapping\footnote{As it will become clearer later, we do not need to give an explicit
expression for this function.} of $\mathbf{x}$ into the space $\mathcal{H},$ $\mathbf{w}=[w_{1},w_{2},...,w_{d}]^{T}$
denotes a $d$-dimensional real vector normal to $H$ and $b$ is
a real parameter such that $\left|b\right|/\left\Vert \mathbf{w}\right\Vert $
is the perpendicular distance of the origin from $H$ (here the operators
$\cdot$ and $\left\Vert \cdot\right\Vert $ denote the scalar product
between two vectors and the Euclidean norm of a vector, respectively).
In practice, the problem of finding $H$ can be formulated as the
(\emph{primal}) optimization problem
\begin{equation}
\underset{\mathbf{w},b,\mathbf{\xi}}{\min}\;\frac{1}{2}\mathbf{w}^{T}\mathbf{w}+C\sum_{i=1}^{N_{train}^{(z)}}\xi_{i}\label{eq:C-SVC}
\end{equation}
subject to the constraints
\begin{equation}
\xi_{i}>0\label{eq:C-SVC-constr}
\end{equation}
and 
\begin{equation}
z_{i}\left[\mathbf{w}^{T}\phi\left(\mathbf{x}_{i}\right)+b\right]\geq1-\xi_{i},\label{eq:constr_C-SVC}
\end{equation}
with $i=1,2,...,N_{train}^{(z)}$. Here, $\xi_{i}$ is the $i$-th
\emph{slack variable} (the slack variables are collected in the vector
$\mathbf{\mathrm{\xi}}\triangleq[\xi_{1},\xi_{2},...,\xi_{N_{train}^{(z)}}]^{T}$)
and $C$ is a \emph{regularization parameter} to be chosen by the
user. It is important to point out that: a) the slack variables are
included in the considered problem to account for the presence of
the above mentioned classification errors (in practice, the occurrence
of a specific error is associated with the corresponding slack variable
exceeding unity, so that the sum $\sum_{i=1}^{N_{train}^{(z)}}\xi_{i}$
appearing in (\ref{eq:C-SVC}) represents an upper bound on the overall
number of these errors); b) consequently, the parameter $C$ allows
the SVM user to control the weight of these errors in the objective
function of (\ref{eq:C-SVC}). The constrained optimization problem
(\ref{eq:C-SVC})-(\ref{eq:constr_C-SVC}) can be easily reformulated
as a \emph{Lagrangian problem}; it is not difficult to prove that
this leads to the \emph{dual problem}
\begin{equation}
\underset{\mathbf{\alpha}}{\min}\;\frac{1}{2}\mathbf{\alpha}^{T}\mathbf{Q}\mathbf{\mathbf{\alpha}}-\mathbf{e}^{T}\mathbf{\mathbf{\alpha}}\label{eq:C-SVC-1}
\end{equation}
subject to
\begin{equation}
\mathbf{z}^{T}\mathbf{\mathbf{\alpha}}=0\label{eq:constr_C-SVC-2}
\end{equation}
and 
\begin{equation}
0\leq\alpha_{i}\leq C,\label{eq:constr_C-SVC-1}
\end{equation}
with $i=1,2,...,N_{train}^{(z)}$. Here, $\mathbf{z}\triangleq[z_{1},z_{2},...,z_{N_{train}^{(z)}}]^{T}$,
$\mathbf{\alpha}\triangleq[\alpha_{1},\alpha_{2},...,\alpha_{N_{train}^{(z)}}]^{T}$
is the vector collecting the (positive) Lagrange multipliers, $\mathbf{e}=[1,1,...,1]^{T}$
is the $N_{train}^{(z)}$-dimensional vector of all ones, $\mathbf{Q}=[Q_{i,j}]$
is a $N_{train}^{(z)}\times N_{train}^{(z)}$ positive semidefinite
matrix with $Q_{i,j}=z_{i}z_{j}K\left(\mathbf{x}_{i},\mathbf{x}_{j}\right)$
and $K\left(\mathbf{x},\mathbf{\hat{x}}\right)\triangleq\phi^{T}\left(\mathbf{x}\right)\phi\left(\mathbf{\hat{x}}\right)$
is the so called \emph{kernel function}, which, as it will become
clearer in Paragraph \ref{sub:training}, contains a single parameter
(denoted $\gamma$) to be optimized. Solving the convex quadratic
optimization problem (\ref{eq:C-SVC-1})-(\ref{eq:constr_C-SVC-1})
yields the vector $\mathbf{\alpha}_{o}=[\alpha_{o,1},\alpha_{o,2},...,\alpha_{o,l}]^{T}$
of Lagrange multipliers, from which the optimal value
\begin{equation}
\mathbf{w}_{o}=\sum_{i=1}^{N_{train}^{(z)}}z_{i}\alpha_{o,i}\phi\left(\mathbf{x}_{i}\right)\label{eq:optimal_w_class}
\end{equation}
of $\mathbf{w}$ and the optimal value $b_{0}$ of $b$ can be evaluated.
Given $\mathbf{w}_{o}$ and $b_{0}$, the \emph{decision function}
\begin{equation}
\begin{array}{c}
d\left(\mathbf{x}\right)\triangleq\mathrm{sgn}\left(\mathbf{w}_{o}^{T}\phi\left(\mathbf{x}\right)+b_{o}\right)\\
=\mathrm{sgn}\left(\sum_{i=1}^{N_{train}^{(z)}}z_{i}\alpha_{o,i}K\left(\mathbf{x}_{i},\mathbf{x}\right)+b_{o}\right),
\end{array}\label{eq:decision_function}
\end{equation}
can be computed. This function is employed to classify an arbitrary
feature vector $\mathbf{x}$ not included in the training set; in
fact, in the considered problem this vector is assigned (not assigned)
to the coverage area if $d\left(\mathbf{x}\right)=1$ ($d\left(\mathbf{x}\right)=-1$).
In particular, the function $d\left(\mathbf{x}\right)$ (\ref{eq:decision_function})
can be evaluated for all the locations which the test set $S_{test}^{(z)}$
refers to in order to assess the classification accuracy. Note that
this accuracy is unavoidably influenced by the two parameters $C$
and $\gamma$ appearing in the objective function of (\ref{eq:C-SVC})
and in the kernel function, respectively. For this reason, a \emph{cross-validation}
procedure is usually needed to adjust these parameters for optimizing
classification performance (e.g., see \cite[Par. 3.2]{Hsu} for a
detailed analysis of this problem), as will be discussed in some detail
in Paragraph \ref{sub:Param-opt}.

The classification procedure described above deserves also the following
comments:
\begin{enumerate}
\item In generating the set\emph{ }$S^{(y)}$, the experimental data collected
in a specific measurement campaign undergo a pseudorandom permutation
in order to avoid that the data collected in $S_{train}^{(z)}$ and
those included in $S_{test}^{(z)}$ refer to disjoint portions of
the area in which such measurements have been collected. Ignoring
this simple rule might entail a significant loss in the \emph{generalization
capability} of our classification and regression methods.
\item In our work, unless explicitly stated, the training set $S_{train}^{(z)}$
contains eighty percent of the data collected in $S^{(z)}$. 
\item As it will become clearer later, the decision function \emph{$d\left(\mathbf{x}\right)$}
(\ref{eq:decision_function}) can be exploited in the regression procedure
described below. 
\end{enumerate}
The other part of our approach is represented by a regression procedure
which generates a prediction of the RSS map inside the coverage area.
From a mathematical viewpoint, such a procedure generates an hyperplane
in the space of features; however, generally speaking, this hyperplane
is different from that produced by our classification method and is
employed for predicting the field strength at unknown locations. Similarly
as the classification procedure, the adopted procedure requires a
\emph{training set} $S_{train}^{(m)}$ and a \emph{test set} $S_{test}^{(m)}$,
which result from partitioning $S^{(m)}$. However, unlike classification,
the procedure for generating these consists of the following steps: 
\begin{itemize}
\item \emph{Step} 1 - $S_{train}^{(m)}$ is obtained from the set $S_{train}\triangleq\{(\mathbf{x}_{i},m_{i}),i=1,2,...,N_{train}^{(z)}\}$
(in one-to-one correspondence with $S_{train}^{(z)}$) by discarding
all the data associated with a value of their index $i$ such that
$z_{i}=-1$ (in other words, the data associated with those locations
at which the transmitted signal is not correctly received are ignored). 
\item \emph{Step} 2 - Similarly as $S_{train}^{(m)}$, $S_{test}^{(m)}$
is obtained from the set $S_{test}\triangleq\{(\mathbf{x}_{i},y_{i}),i=N_{train}^{(z)},N_{train}^{(z)}+1,...,N_{train}^{(z)}+N_{test}^{(z)}\}$
(in one-to-one correspondence with $S_{test}^{(z)}$) by discarding
all the data associated with a value of their index $i$ such that
$z_{i}=-1$. 
\end{itemize}
An additional procedure for further reducing the amount of data included
in $S_{train}^{(m)}$ is carried out within step 2 when our classification
and regression models are trained on the basis of a set of measurements
that have not been acquired in the same environment as that for which
the identification of the coverage area and the prediction are accomplished;
this procedure involves the function \emph{$d\left(\mathbf{x}\right)$}
(\ref{eq:decision_function}) and is described in Paragraph \ref{sub:software-tool}.
Generally speaking, our set generation method results in a training
set $S_{train}^{(m)}\triangleq\{(\mathbf{\tilde{x}}_{i},\tilde{m}_{i}),i=1,2,...,N_{train}^{(m)}\}$
and in a test set $S_{test}\triangleq\{(\mathbf{\tilde{x}}_{i},\tilde{m}_{i}),i=N_{train}^{(m)}+1,N_{train}^{(m)}+2,...,N_{train}^{(m)}+N_{test}^{(m)}\}$;
here, the notation $\left(\mathbf{\tilde{x}}_{i},\tilde{m}_{i}\right)$
has been adopted in place of $\left(\mathbf{x}_{i},m_{i}\right)$
since a renumbering of the remaining data is done at the end of step
2 within each set; moreover $N_{train}^{(m)}$ and $N_{test}^{(m)}$
denote the number of elements of $S_{train}^{(m)}$ and $S_{test}^{(m)}$,
respectively. Then, $S_{train}^{(m)}$ is processed by a $\epsilon$
- s\emph{upport vector regression} ($\epsilon$-SVR) algorithm \cite{Vapnik_1998}.
This requires solving the (primal) optimization problem
\begin{equation}
\underset{\mathbf{\tilde{w}},\tilde{b},\mathbf{\tilde{\xi}},\mathbf{\tilde{\xi}}^{*},\epsilon}{\min}\;\frac{1}{2}\mathbf{\tilde{w}}^{T}\mathbf{\tilde{w}}+\tilde{C}\sum_{i=1}^{N_{train}^{(y)}}\tilde{\xi}_{i}+\tilde{C}\sum_{i=1}^{N_{train}^{(y)}}\tilde{\xi}_{i}^{*}\label{eq:C-SVC-2}
\end{equation}
subject to 
\begin{equation}
\tilde{\xi}_{i},\tilde{\xi}_{i}^{*}\geq0,
\end{equation}
 
\begin{equation}
\mathbf{w}^{T}\phi\left(\mathbf{\tilde{x}}_{i}\right)+\tilde{b}-\tilde{m}{}_{i}\leq\epsilon+\tilde{\xi}_{i}\label{eq:constr_C-SVR1}
\end{equation}
and
\begin{equation}
\tilde{m}_{i}-\mathbf{w}^{T}\phi\left(\mathbf{\tilde{x}}_{i}\right)-\tilde{b}\leq\epsilon+\tilde{\xi}_{i}^{*},\label{eq:constr_C-SVR1-1}
\end{equation}
with $i=1,2,...,N_{train}^{(m)}$. The meaning of most parameters
and vectors appearing in the new problem is similar to that illustrated
for the related problem (\ref{eq:C-SVC})-(\ref{eq:constr_C-SVC});
note, in particular, that $\tilde{\xi}_{i}^{*}$, with $i=1,2,...,N_{train}^{(m)}$,
is a slack variable appearing in the new constraint (\ref{eq:constr_C-SVR1-1}),
$\mathbf{\mathrm{\xi}}\triangleq[\xi_{1},\xi_{2},...,\xi_{l}]^{T}$
and that $\tilde{C}>0$ is a regularization parameter modifying the
weight of the errors appearing in the regression procedure. However,
in this case, the constrains (\ref{eq:constr_C-SVR1}) and (\ref{eq:constr_C-SVR1-1})
contain the additional real parameter $\epsilon>0$, which represents
the width of the called $\epsilon$-\emph{insensitive tube} (i.e.,
a measure of the precision of regression). In our work $\epsilon=3$
has been selected on the basis of: a) the measurement noise intensity
(see (\ref{nonlinearmodel})) estimated in our measurement campaign;
b) a trial and error procedure adopted to assess the sensitivity of
the $\epsilon$-SVR procedure to the parameter $\epsilon$. Once the
value of $\epsilon$ is selected, similarly as the classification
technique, the only parameters whose values need to be properly adjusted
in the $\epsilon$-SVR technique are $\tilde{C}$ and $\tilde{\gamma}$
(the last one is contained in the kernel function).

Reformulating the optimization problem (\ref{eq:C-SVC-2})-(\ref{eq:constr_C-SVR1-1})
as a Lagrangian problem leads easily to the \emph{dual problem} 
\begin{equation}
\begin{array}{c}
\underset{\mathbf{\tilde{\alpha}},\mathbf{\mathbf{\alpha^{*}}}}{\min}\;\frac{1}{2}\left(\mathbf{\tilde{\alpha}}-\tilde{\alpha}^{*}\right)^{T}\mathbf{\mathbf{\tilde{Q}}}\left(\mathbf{\tilde{\alpha}}-\tilde{\alpha}^{*}\right)+\epsilon\sum_{i=1}^{N_{train}^{(y)}}\left(\tilde{\alpha}_{i}+\tilde{\alpha}_{i}^{*}\right)\\
+\sum_{i=1}^{N_{train}^{(y)}}\tilde{m}{}_{i}\left(\tilde{\alpha}_{i}-\tilde{\alpha}_{i}^{*}\right)
\end{array}\label{eq:C-SVC-2-2}
\end{equation}
subject to
\begin{equation}
\mathbf{\tilde{e}}^{T}\mathbf{\mathbf{\left(\mathbf{\tilde{\alpha}}-\tilde{\alpha}^{*}\right)}}=0
\end{equation}
and 
\begin{equation}
0\leq\tilde{\alpha}_{i},\tilde{\alpha}_{i}^{*}\leq\tilde{C,}\label{eq:constr_C-SVC-1-1}
\end{equation}
with $i=1,2,...,N_{train}^{(m)}$. Here, $\tilde{\mathbf{e}}=[1,1,...,1]^{T}$
is the $N_{train}^{(m)}$-dimensional vector of all ones, $\mathbf{\tilde{\alpha}}=[\tilde{\alpha}_{1},\tilde{\alpha}_{2},...,\tilde{\alpha}_{N_{train}^{(m)}}]^{T}$
and $\mathbf{\tilde{\alpha}^{*}}=[\tilde{\alpha}_{1}^{*},\tilde{\alpha}_{2}^{*},...,\tilde{\alpha}_{N_{train}^{(m)}}^{*}]^{T}$
are $N_{train}^{(m)}$-dimensional vectors collecting the (positive)
Lagrange multipliers, and $\mathbf{\tilde{Q}}=[\tilde{Q}_{i,j}]$
is a $N_{train}^{(m)}\times N_{train}^{(m)}$ positive semidefinite
matrix, with $\tilde{Q}_{i,j}=K\left(\mathbf{x}_{i},\mathbf{x}_{j}\right)$
(note that in our approach the mapping $\mathbf{\phi}\left(\mathbf{x}\right)$
and the kernel function $K\left(\mathbf{x},\mathbf{\hat{x}}\right)$
are the same as those employed in the classification algorithm described
above). Solving the problem (\ref{eq:C-SVC-2-2})-(\ref{eq:constr_C-SVC-1-1})
produces the solutions $\mathbf{\tilde{\alpha}}_{o}=[\tilde{\alpha}_{o,1},\tilde{\alpha}_{o,2},...,\tilde{\alpha}_{o,N_{train}^{(m)}}]^{T}$
and $\mathbf{\tilde{\alpha}}_{o}^{*}=[\tilde{\alpha}_{o,1}^{*},\tilde{\alpha}_{o,2}^{*},...,\tilde{\alpha}_{o,N_{train}^{(m)}}^{*}]^{T}$
for the vectors $\mathbf{\tilde{\alpha}}$ and $\mathbf{\tilde{\alpha}}^{*}$,
respectively, and $\tilde{b}_{o}$ for $\tilde{b}$; this allow us
to evaluate the \emph{predictive function} 
\begin{equation}
p\left(\mathbf{x}\right)\triangleq\sum_{i=1}^{N_{train}^{(m)}}\left(\tilde{\alpha}_{i}^{*}-\tilde{\alpha}_{i}\right)K\left(\mathbf{\tilde{x}}_{i},\mathbf{x}\right)+\tilde{b}_{o},\label{eq:predictive_function}
\end{equation}
which represents the solution to our regression problem. 

Finally, it is worth pointing out that:
\begin{itemize}
\item The accuracy of the field strength prediction provided by the predictive
function $p\left(\mathbf{x}\right)$ (\ref{eq:predictive_function})
is influenced not only by the data contained in the training set $S_{train}^{(m)}$
, but also by our specific choice for the values of the parameters
$\tilde{C}$ and $\tilde{\gamma}$. The optimization of these parameters
is accomplished on the basis of the test set $S_{test}^{(m)}$ and
of a specific performance measure, and involves a cross-validation
procedure; this is discussed in more detail in Paragraph \ref{sub:Param-opt}.
\item Our approach exploits both a classification technique for delimiting
the coverage area and a regression technique for estimating the field
strength inside it. In previous work investigating the application
of SVM to field strength prediction (see \cite{Lagan=0000E0_2009},
\cite{Piacentini_2010} and \cite{Robs14}), the use of SVM-based
regression only has been proposed, so that the system coverage area
is generated as a by-product of the adopted regression method. However,
our numerical results have evidenced that the last approach may result
in an appreciably less accurate estimate of the coverage area with
respect to the method we propose in this manuscript. 
\end{itemize}

\section{Measurement tools and campaigns\label{sec:Instr}}

In this Section a short description of the measurement tools and of
the hardware set-up adopted in our measurement campaigns is provided.
Then, some relevant information about the scenarios considered in
such campaigns and various specific guidelines followed in acquiring
our measurements are illustrated.

\subsection{Measurement tools and set-up\label{sub:Datacoll}}

\begin{figure}
\centering{}\includegraphics[width=0.85\columnwidth]{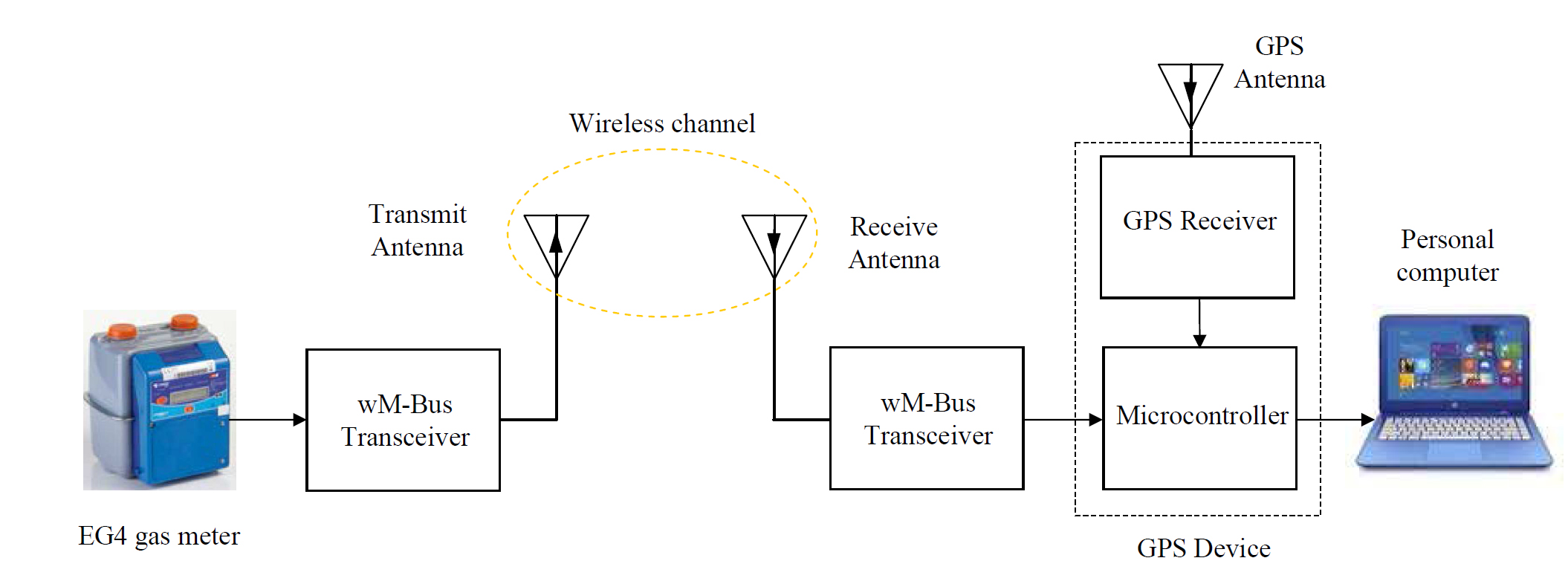}\caption{Block diagram of the hardware tools employed in our measurement campaign.\label{fig:Sistema-radio-trasmettitore-rice}}
\end{figure}

As already mentioned in Paragraph \ref{sub:Problem-statement}, we
are interested in measuring the outdoor propagation losses in a wireless
star network consisting of a data concentrator serving multiple gas
meters. A block diagram of the hardware tools employed in our measurement
campaigns is shown in Fig. \ref{fig:Sistema-radio-trasmettitore-rice}.
In practice, to ease data acquisition, a commercial gas meter, equipped
with a 169 MHz transceiver and placed in a \emph{fixed} outdoor location
(of potential interest for the installation of a data concentrator),
has been used for accomplishing wireless data transmission; on the
other hand, a data acquisition hardware, including another 169 MHz
transceiver, a GPS receiver and a portable PC, has been placed on
a car moving at a limited and approximeted constant speed in a wide
area surrounding the transmitter. Some relevant technical information
about the transmitter, the data acquisition device and its output
information are provided below.

\begin{figure}
\centering{}
\includegraphics[width=0.45\textwidth]{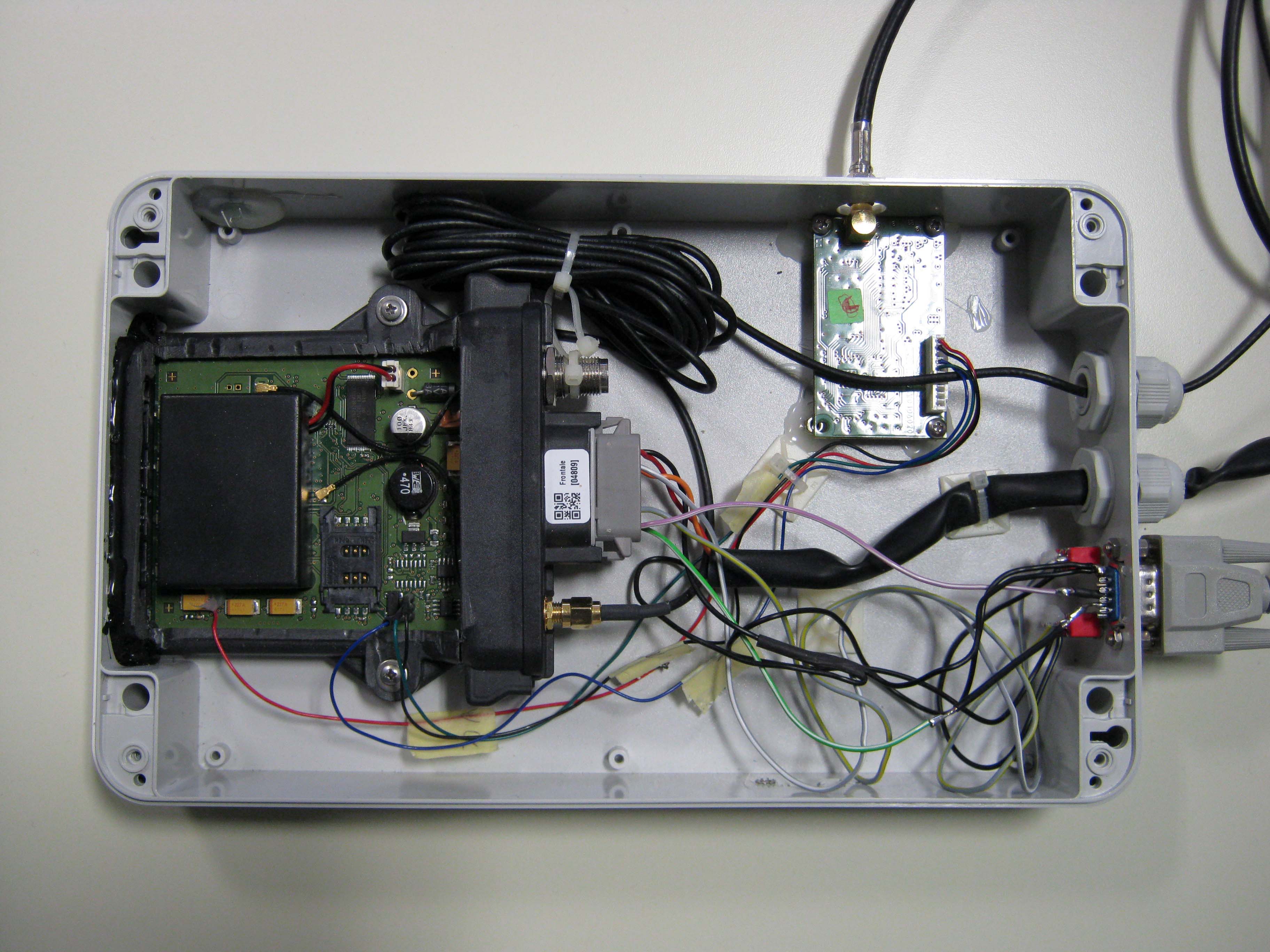}
\caption{Data acquisition device employed in our measurement campaigns (the antenna and the portable computer connected to this device are not visible in this photograph).
\label{fig:Ricevitore}}
\end{figure}

\begin{figure}
\centering{}\includegraphics[width=0.45\textwidth]{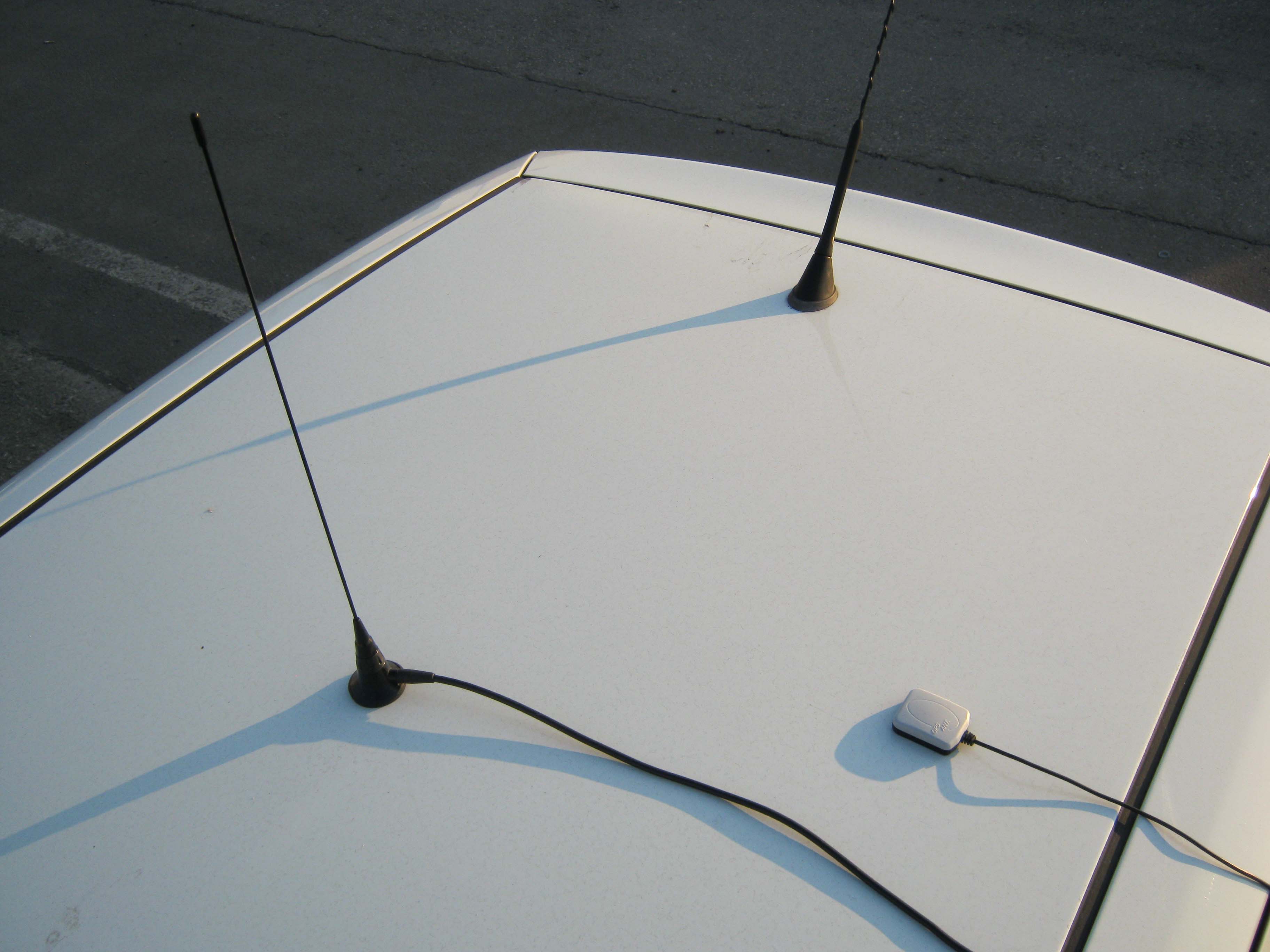}
\caption{Sirio SKB antenna (on the left) and GPS antenna (on the right) employed by our data acquisition hardware installed on the roof of the car
employed in our measurement campaign.
\label{fig:Installazione-su-auto}}
\end{figure}

\begin{enumerate}
\item \emph{Transmitter}- This device is made of a EG4 gas meter \cite{meter_italia}
manufactured by the Italian company Meter Italia (which is a subsidiary
company of CPL Concordia) and a wireless M-Bus transceiver, operating
according to the mode N1c of EN 13757-4:2013 \cite{EN13757-4:2013}
and transmitting\footnote{Unluckily, technical information about the transmit antenna employed
in this gas meter cannot be made public.} a new meter reading every 3 s. In practice, a new data packet, consisting
of 46 bytes (35 of which form the payload), is radiated every 3 s;
each packet contains the gas meter address and a fictitious gas reading
(since the gas meter is not connected to a gas network), and its transmission
lasts about $0.153$s (since the bit rate for 2a channel of the N1c
mode is equal to 2.4 kbps); one of the following power levels can
be selected for data transmission: 21 dBm, 24 dBm, 27 dBm and 30 dBm.
\item \emph{Data acquisition hardware} - This device has been explicitly
designed and implemented for our measurement campaigns (see Fig. \ref{fig:Ricevitore}).
It consists of: a) a receiver board based on a wireless M-Bus transceiver
RC1701HP-MBUS, manufactured by Radiocrafts \cite{Radiocrafts} and
characterized by a nominal sensitivity\footnote{Our transceiver is unable to detect the transmitted signal if its
strength is below this threshold. For this reason, as already mentioned
above, a power level of $P_{NC}=-120\,\mbox{dBm}$ has been conventionally
selected to denote the lack of radio coverage at a specific location.} equal to -119 dBm; b) a Sirio SKB antenna \cite{Sirio} (i.e., a
$\lambda/4$ omnidirectional antenna for the VHF band) feeding the
transceiver and installed on the roof of a car (the roof behaves like
a ground plane; see Fig. \ref{fig:Installazione-su-auto}); c) an
AM50 GPS receiver manufactured by GeoShack \cite{GeoShack} and equipped
with an external antenna; d) a portable computer. The data acquisition
hardware operates as follows. A new estimate of the strength of the
signal received from the gas meter of interest, together with the
meter address, is sent to a microcontroller embedded in the GPS module
every 3 s; these information are tagged by an estimate of the car
position provided by the GPS receiver and then sent to a portable
computer (through a serial port), where they are displayed in real
time and stored in a text file. 
\item \emph{Available information} - The data available on the personal
computer include the following information: date, timing, longitude
and latitude (expressed in decimal degrees), altitude (expressed in
m), car speed (expressed in m/s), car direction, number of GPS satellites,
meter address and an RSS \emph{indication} (RSSI, expressed in dBm)
averaged over the considered data packet.
\end{enumerate}

\subsection{Measurement campaigns: selected scenarios and guidelines followed
in data acquisition\label{sub:meascamp}}

Our measurement campaigns have mainly concerned densely populated
areas of cities and villages in northern Italy (i.e., \emph{urban}
and \emph{suburban} scenarios), because of their relevance in the
deployment of new infrastructures for gas metering. The considered
areas have partitioned into two classes, the first one (class \#1)
including the villages and the cities located in the Po valley\footnote{This is a large and flat area delimited by the Italian Alps, the Adriatic
sea and the Tuscan-Emilian Apennine.} (i.e., characterized by a \emph{flat} terrain), the second one (class
\#2) encompassing villages located on the Italian Alps or on the Tuscan-Emilian
Apennine (i.e., characterized by a \emph{hilly} or \emph{mountainous}
terrain); further details about the selected areas are given in Section
\ref{sec:Numerical-and-experimental}. As it will become clearer in
Section \ref{sub:features}, our classification of the selected geographical
areas is based on the fact that, when the points at which measurements
are collected in a given area are characterized by significantly different
altitudes, a couple of specific features, which are not taken into
consideration for flat terrains, are employed in SVM classification
and regression; vice versa, feature extraction for flat terrains entails
the evaluation of two features which are ignored in the other case. 

In our work the following two specific tasks have been carried out
before starting a new measurement campaign in each of the selected
urban scenarios:
\begin{enumerate}
\item \emph{Partitioning of the considered scenario into districts} - Large
urban areas have been usually partitioned into districts characterised
by similar propagation conditions. This step is motivated by our interest
in comparing field strength models referring to different cities or
villages, but whose data have been collected in districts exhibiting
a significant degree of similarity in terms of street widths, building
densities, building heights, building age and vegetation.
\item \emph{Selection of transmitter locations} - In each district favorite
locations for our transmitter have been identified by looking for
proper positions of data concentrators. Note that, first of all, such
positions should be easily accessible to ease the installation of
the transmitter; moreover, when possible, they should be sufficiently
high to ensure a large coverage area within the considered district.
Unluckily, we were often unable to access relevant sites characterized
by a significant height, so that the height of the transmitter usually
ranged from 1,5 m to 1,8 m (a few high positions, available on different
buildings, were available in the city of Modena and in the village
of Rocca Malatina only). However, it should be expected that, in any
future infrastructure for gas metering, heterogeneous sites, characterized
by significantly different heights, will be available.
\end{enumerate}
After this preliminary work, the transmitter has been installed in
one of the selected locations and RSS data have been acquired in all
the roads of the considered district and its neighborhoods travelling
at an almost constant speed (25 km/h); this has allowed us to achieve
a uniform spatial sampling of the considered area and to maximise
the diversity of the acquired data. As far as the last point is concerned,
it is important to mention that, in each measurement campaign, we
have tried to collect similar quantities of measurements for the coverage
area and for the area outside it, since, as evidenced by our computer
simulations, this improves the accuracy of our classification method.
Note that, generally speaking, the availability of balanced quantities
of data associated with the presence of radio coverage and with the
lack of it helps a classification algorithm to learn to discriminate
between these two different conditions. At the same time, a sufficiently
large number of points acquired in the coverage area improves the
prediction capability of the employed regression algorithm.

\section{Data processing and software tools\label{sec:Dataproc}}

As explained in the previous Section, each measurement campaign has
resulted in a set of RSS measurements referring to a specific district
of a given town or village; each measurement is tagged by the GPS
coordinates of the location at which it has been acquired. In this
Section we provide some additional details about the processing techniques
employed to predict a coverage area and a RSS map on the basis of
the available measurements. In particular, we first describe the procedure
we adopted to extract a set of features from each of the considered
locations. Then, we take into consideration SVM processing again,
and discuss some specific choices we made in implementing it. Finally,
we comment on our software implementation of the proposed method and
briefly illustrate the potentialities of the SVM-based software tool
we developed for radio planning in the 169 MHz band.

\subsection{Extraction of significant features\label{sub:features}}

\begin{figure}
\centering{}\includegraphics[width=1\columnwidth]{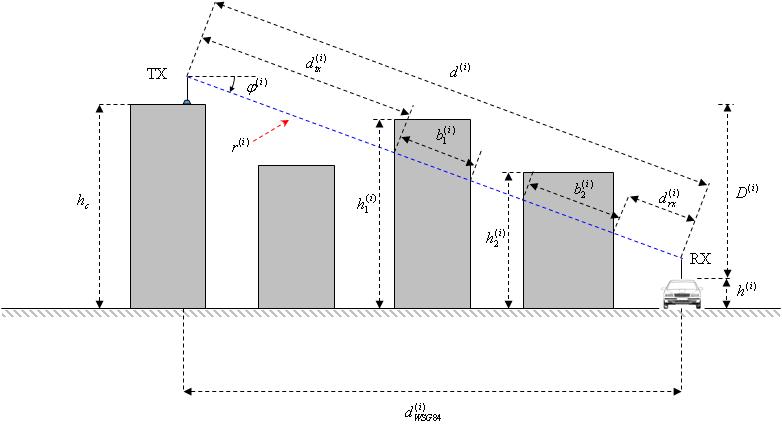}\caption{Schematic representation of an urban scenario in which the line-of-sight
wireless link between a transmitter and a receiver is blocked by multiple
buildings. \label{fig:building_geometry}}
\end{figure}

In this Section we assume that, for a given location of the transmitter,
$N$ measurements have been collected at $N$ distinct and known locations
in its neighborhoods. Before starting SVM classification, three groups
of geometric features are evaluated for the $i$-th location (with
$i=1,2,...,N$) on the basis of: a) the positions of the transmitter
and the receiver, b) a 3D geometrical representation of the considered
propagation environment (in practice, a digital map). The first group,
which is computed on the basis of the positions of the transmitter
and the receiver only, consists of the following three features (see
Fig. \ref{fig:building_geometry}):
\begin{enumerate}
\item \emph{Transmitter-receiver differential height} - This feature is
defined as 
\begin{equation}
D^{(i)}\triangleq h_{c}-h^{(i)},\label{eq:vertical-drop}
\end{equation}
where $h_{c}$ denotes the altitude of the \emph{transmitter} (TX)
and $h^{(i)}$ the altitude of the $i$-th position for the \emph{receiver}
(RX) (each altitude is measured with respect to the support of the
corresponding antenna).
\item \emph{Transmitter-receiver distance} - The distance between the transmitter
and the $i$-th position is evaluated as 
\begin{equation}
d^{(i)}=\sqrt{(D^{(i)})^{2}+(d_{WGS84}^{(i)})^{2}},\label{eq:distance}
\end{equation}
where $D^{(i)}$ is given by (\ref{eq:vertical-drop}) and $d_{WGS84}^{(i)}$
represents the distance between the considered positions evaluated
on the basis of their WGS-84 coordinates\footnote{In practice, the Vincenty's inverse method \cite{Vincenty} is used
to compute the geographical distance $d_{WGS84}^{(i)}$ between the
TX and RX locations on the basis of their GPS coordinates. }. Note that $d^{(i)}$ represents the length of the segment $r^{(i)}$
connecting the transmitter to the receiver, as illustrated in Fig.
\ref{fig:building_geometry}.
\item \emph{Angular deviation} - This feature (denoted $\varphi^{(i)}$
in Fig. \ref{fig:building_geometry}) represents the elevation angle
of the RX antenna measured with respect to the TX antenna; note that
in our measurements the employed antennas exhibit an omnidirectional
pattern on their azimuth planes and that they are never tilted. The
adoption of this feature is motivated by the fact that such antennas
do not have an omnidirectional pattern on the elevation plane, so
that an angular deviation different from zero may have a significant
impact on propagation losses.
\end{enumerate}
The second group of features depends not only on the positions of
the transmitter and the receiver, but also on the positions and geometry
of the buildings preventing \emph{line of sight} (LOS) propagation,
as illustrated in Fig. \ref{fig:building_geometry}; in particular,
these features aim at capturing some relevant properties of key obstructions.
In the following, it is assumed that at the $i$-th position of the
receiver (with $i=1,2,...,N$) the view of the transmitter is obstructed
by $K^{(i)}$ buildings, whose heights are denoted $\{h_{l}^{(i)},l=1,2,...,K^{(i)}\}$
(note that $K^{(i)}=2$ in Fig. \ref{fig:building_geometry}). Moreover,
the length of the portion of the segment $r^{(i)}$ intersecting the
$l$-th building is denoted $b_{l}^{(i)}$. Given these definitions,
the following five features are evaluated at the $i$-th location:
\begin{enumerate}
\item \emph{Maximum height of blocking buildings} - This feature is defined
as 
\begin{equation}
h_{max}^{(i)}\triangleq\underset{l}{\max\,}h_{l}^{(i)}\label{eq:h_max}
\end{equation}
and provides information about the maximum degree of the vertical
obstruction along the $r^{(i)}$direction.
\item \emph{Average height of blocking buildings} - This feature is defined
as
\begin{equation}
h_{av}^{(i)}\triangleq\frac{\sum_{l=1}^{K^{(i)}}h_{l}^{(i)}}{K^{(i)}}\label{eq:h_med}
\end{equation}
and provides information about the average degree of the vertical
obstruction encountered along the $r^{(i)}$ direction.
\item \emph{Portion through buildings} (PTB) - This feature is evaluated
on the basis of the formula 
\begin{equation}
PTB^{(i)}\triangleq\frac{{\textstyle \sum_{l=1}^{K^{(i)}}b_{j}^{(i)}}}{d^{(i)}}\label{eq:ptb}
\end{equation}
and provides information about �the portion of $r^{(i)}$ included
in the blocking buildings. 
\item \emph{Distance between transmitter and its closest building} - This
feature, denoted $d_{tx}^{(i)}$ in Fig. \ref{fig:building_geometry},
represents the distance between the transmitter and its closest building,
measured along the $r^{(i)}$ direction. Its use is motivated by the
fact that the building closest to the transmitter is expected to have
a significant influence on the attenuation of the transmitted signal.
\item \emph{Distance between receiver and its closest building} - This feature,
denoted $d_{rx}^{(i)}$ in Fig. \ref{fig:building_geometry}, represents
the distance between the receiver and its closest building, measured
along the $r^{(i)}$ direction. This feature is the dual of the previous
one.
\end{enumerate}
The last group consists of a single feature, namely the so-called
\emph{portion through ground} (PTG), which plays a fundamental role
in the presence of an uneven terrain. The definition of this quantity
is very similar to that of $PTB^{(i)}$ (\ref{eq:ptb}), since it
refers to the case in which the view of the transmitter is obstructed
by ground at the receive side; in this case we assume that, generally
speaking, the segment $r^{(i)}$ connecting the transmitter to the
receiver intersects ground $N^{(i)}$ times and that $a_{l}^{(i)}$
denotes the length of its $l$-th intersection (with $l=1,2,...,N^{(i)}$).
Then, the PTG at the $i$-th location is defined as 

\begin{equation}
PTG^{(i)}\triangleq\frac{{\textstyle \sum_{l=1}^{N^{(i)}}a_{l}^{(i)}}}{d^{(i)}},\label{eq:ptg}
\end{equation}
with $i=1,2,...,N$.

Independently of the class the analysed area belongs to, in our data
processing procedure the following five features have been computed:
transmitter-receiver distance, angular deviation, maximum height of
blocking buildings, average height of blocking buildings and PTB.
Moreover, the following two features have been computed for class
\#1 only: distance between\emph{ }transmitter and its closest building,
and distance between receiver and its closest building\emph{. }On
the contrary, the remaining two features (namely, transmitter-receiver
differential height and PTG) have been computed for class \#2 only.
Therefore, whatever the class of the analysed area, the feature extraction
procedure generates a set $S^{(\tilde{x})}\triangleq\mathbf{\{\tilde{x}}_{l},\,l=1,2,...,N\}$
of seven dimensional vectors (here, $\mathbf{\tilde{x}}_{l}=[\tilde{x}_{l}^{(1)},\tilde{x}_{l}^{(2)},...,\tilde{x}_{,l}^{(m)}]^{T}$
with $m=7$), which, as discussed in the following Paragraph, are
pre-processed before feeding the classification and regression algorithms
illustrated in Section \ref{sub:Theory}.

Finally, it is worth mentioning that:
\begin{enumerate}
\item All the features defined above have a geometric meaning. However,
it is well known that, generally speaking, radio propagation in a
given environment is influenced by a number of factors among which
an important role may be played by atmospheric conditions. Fluctuations
of a few dBm were observed in the received power when the transmitter
and the receiver were held at fixed locations for some days; however,
we were unable to separate the contribution originating from atmospheric
factors (e.g., humidity and temperature) from those due to other environmental
changes occurring in the neighborhoods (e.g., reflection and blocking
due to vehicles, pedestrians and other objects on close roads). 
\item The selection of the features employed in our work has been inspired,
in part, by previous work and, in particular, by \cite{Piacentini_2010}.
Moreover, extensive computer simulations have been run to experimentally
assess the relevance of each feature, i.e. its impact on the accuracy
of the considered SVM-based methods. Our numerical results have evidenced
that a good accuracy can be achieved if the sets of features defined
above are adopted. Note that these sets are small so that \emph{dimensionality
reduction techniques} \cite{Bishop} are not needed. On the contrary,
a larger number of features is evaluated in \cite{Piacentini_2010},
since the segment $r^{(i)}$ connecting the transmitter to the receiver
(see Fig. \ref{fig:building_geometry}) is partitioned into multiple
intervals, and three distinct features are evaluated for each interval;
for this reasons, \emph{principal component analysis} and \emph{nonlinear}
\emph{principal component analysis} are used in that case to reduce
the dimensionality of the feature space before SVM-based processing
in order to prevent the well known \emph{overfitting} phenomenon.
\end{enumerate}

\subsection{Data Pre-Processing and SVM Processing\label{sub:training}}

The extraction of the features defined in the previous Paragraph is
followed by data pre-processing (in particular, data scaling) and
SVM processing (whose mathematical details have been illustrated in
Section \ref{sub:Theory}). In the following we describe the method
we adopted for data scaling and then provide some additional details
about the SVM processing we implemented. 
\begin{itemize}
\item \emph{Data scaling} - Data scaling is usually employed in SVM applications
when the components of feature vectors are characterized by substantially
different ranges and, in particular, some of them may take on large
values. In fact, when this occurs, large components may hide the contribution
of the remaining ones in the evaluation of kernel values\footnote{Note that typical kernels depend on feature vectors through their
inner products.}; this unavoidably entails various numerical problems, resulting in
a loss of accuracy. In the considered problem the ranges of some components
of the feature vector $\mathbf{\tilde{x}}_{l}=[\tilde{x}_{l}^{(1)},\tilde{x}_{l}^{(2)},...,\tilde{x}_{l}^{(m)}]^{T}$
(with $m=7$ and $l=1,2,...,N$) are substantially different (see
(\ref{eq:vertical-drop})-(\ref{eq:ptg})), so that scaling is indispensable.
For this reason, in our work the vector $\mathbf{\tilde{x}}_{l}$
is mapped into the scaled feature vector $\mathbf{x}_{l}=[x_{l}^{(1)},x_{l}^{(2)},...,x_{l}^{(m)}]^{T}$for
any $l$; the adopted mapping rule can be expressed as
\begin{equation}
x_{l}^{(i)}=\frac{\tilde{x}_{l}^{(i)}-\mu_{i}}{\delta_{i}}
\end{equation}
with $i=1,2,...,m$, where $\mu_{i}$ and $\delta_{i}$ denote the
mean value and the standard deviation, respectively, of the $i$-th
feature. This leads to generating the new set of feature vectors $S^{(x)}\triangleq\{\mathbf{x}_{l},\,l=1,2,...,N\}$,
which feeds our SVM-based classification and regression techniques.
\item \emph{Selection of the kernel function} - In the SVM literature various
options are available for the kernel function; in particular, the
most common options are the linear, polynomial, sigmoid and \emph{radial
basis function} (RBF) kernels (e.g., see \cite[Sec. 1]{Hsu} and \cite[Par. 4.3]{Burges}).
Generally speaking, the RBF kernel
\begin{equation}
K(\mathbf{x},\mathbf{y})=\exp(-\gamma\left\Vert \mathbf{x}-\mathbf{y}\right\Vert ^{2}),
\end{equation}
is the most reasonable choice in a number of problems (here, $\gamma$
is a real positive parameter). In fact, this kernel has a number of
good features, since it can properly handle the cases in which the
relation between class labels and features is nonlinear in classification
problems, it depends on a single hyper-parameter ($\gamma$) and its
use does not usually entail numerical problems \cite[Sec. 3]{Hsu},
\cite{Bishop}. 
\item \emph{Performance metrics} - In our technique the training phases
of SVM-based classification and SVM-based regression are followed
by a validation phase in which the accuracy of the trained models
for classification and for signal strength prediction, respectively,
is assessed. In our work the evaluation of \emph{classification accuracy}
is based on the following performance metrics: a) accuracy; b) full-scale
accuracy; c) percentage of false positives. The \emph{accuracy} (denoted\emph{
$A$} in the following) is defined as
\begin{equation}
A=\frac{N_{cc}}{N_{t}^{(z)}}\cdot100\%,\label{eq:accuracy}
\end{equation}
where $N_{t}^{(z)}$ is the overall number of elements of the test
set $S_{test}^{(z)}$ selected in the classification procedure and
$N_{cc}$ is the number of points belonging to the same set and correctly
classified (in other words, $N_{cc}$ represents the overall number
of true positives and true negatives). A similar definition is adopted
for the \emph{full-scale accuracy} \emph{$A_{fs}$}, which, however,
does not refer to the entire $S_{test}^{(z)}$, but only the portion
of its points characterized by a RSS belonging to the interval {[}-119
dBm, -110 dBm{]} or equal to -120 dBm; note that this parameter has
a significant practical relevance, since it refers to the cases in
which RSS is close to receiver sensitivity (this usually occurs, for
instance, at receiver positions close to the borders of the considered
coverage area). The \emph{percentage of false positives} (denoted
$P_{fp}$ in the following) is defined as the percentage of points
of the test set $S_{test}^{(z)}$ which are erroneously assigned to
the coverage area by the classification algorithm.\\
On the other hand, the evaluation of \emph{regression accuracy} is
based on the evaluation of the \emph{root mean square error} (RMSE),
which is defined as
\begin{equation}
RMSE\triangleq\sqrt{\frac{\sum_{i=N_{train}^{(m)}+1}^{N_{train}^{(m)}+N_{test}^{(m)}}(\hat{m}_{i}-\tilde{m}_{i})^{2}}{N_{test}^{(m)}}},
\end{equation}
where $\hat{m}_{i}$ is the RSS value predicted at the same location
as the measurement $\tilde{m}_{i}\in S_{train}^{(m)}$ and $N_{test}^{(m)}$
is the overall number of elements of the test data set for SVM regression.
\end{itemize}

\subsection{Implementation of the Proposed SVM-Based Method for Radio Planning\label{sub:software-tool}}

A complete software tool based on the proposed method for SVM-based
radio planning has been implemented in MATLAB; this allowed us to
benefit from the availability of various existing packages for this
environment and to easily develop a user friendly \emph{graphical
user interface} (GUI). In the development of this tool the following
two specific problems had to be solved: 
\begin{enumerate}
\item the automatic extraction of the feature vectors for all the locations
at which RSSI data were acquired during each measurement campaign;
\item the implementation of SVM methods.
\end{enumerate}
The first problem has been solved as follows. As already mentioned
above, 3D maps of the considered environments have been downloaded
from public regional databases collecting topographic information
or made available by Italian regional offices for research purposes;
in all the considered cases the available maps were in \emph{shapefile}
(.shp) format. After acquiring the map of the environment of interest,
a preliminary check has been accomplished to assess the quality of
its content (i.e., of the geographical data it contained) and verify
that its format\footnote{Note that our preliminary analysis of the available maps represents
an unavoidable task, because of the lack of a national standard for
open geographical data in Italy; for this reason different Italian
regions may not follow the same guidelines in this field. } was ready to be correctly interpreted by our software tool. This
task has been carried out by means of the open source software QGIS
\cite{QGIS}; in fact, this has allowed us to easily access the map
\emph{layers} referring to: a) 3D representations of buildings; b)
contour lines for the considered area; c) roads. Note that, on the
one hand, building heights and the geographical coordinates of building
corners are needed to evaluate the features $h_{max}^{(i)}$ (\ref{eq:h_max}),
$h_{av}^{(i)}$ (\ref{eq:h_med}), $PTB^{(i)}$ (\ref{eq:ptb}), $d_{tx}^{(i)}$
and $d_{tx}^{(i)}$ (see Fig. \ref{fig:building_geometry}). On the
other hand, the knowledge of contour lines is certainly required in
the evaluation of $PTG^{(i)}$ (\ref{eq:ptg}) for any area belonging
to class \# 2; however, in a hilly/mountainous environment these information
also play a fundamental role in the evaluation of the differential
altitude $D^{(i)}$ (\ref{eq:vertical-drop}), the distance $d^{(i)}$
(\ref{eq:distance}) and the angular deviation $\varphi^{(i)}$ (see
Fig. \ref{fig:building_geometry}). In fact, level curves can be employed
to evaluate the altitude of an arbitrary point by means of a simple
interpolation function, since adjacent curves are usually characterized
by a constant spacing. In principle, as already mentioned in Paragraph
\ref{sub:Problem-statement}, map information about the roads crossing
the considered area are not required. In practice, however, they have
been useful in densely populated areas characterized by high buildings.
In fact, in those cases the acquired GPS positions were often affected
by non negligible errors because of poor reception conditions; when
this occurred, errors have been mitigated projecting each estimated
position onto the central line of the road in which GPS signals have
been acquired.

After properly checking the format and content of the map of the considered
area, map information have been processed to compute the feature vectors
for all the locations at which measurements were acquired. Note that
the implementation of this task has been substantially simplified
by the use of the Mapping Toolbox \cite{Mapping_Toolbox}, which has
allowed us to import the needed geographical data in the Matlab environment.

The second problem, i.e. the implementation of SVM methods, has been
easily solved thanks to the availability of a public library for SVM
methods, called LIBSVM \cite{LIBSVM}. 

In developing our software implementation of the adopted SVM-based
methods (our implementation is called \emph{SVM-based Tool for Radio
Planning}, SVMTRP, in the following) three different \emph{prediction
modes} (PMs) were taken into consideration in order to satisfy all
the needs envisaged for the future radio planning of gas metering
networks. A short description of the aim, input data, data processing
tasks and resulting outputs is provided below for each of these PMs.

\textbf{PM \#1} - \emph{Aim}: this mode allows to assess the accuracy
of the classification and regression techniques implemented in our
study; it should be considered as a tool for assessing the prediction
capability of the employed methods. 

\emph{Input data}: geographical data of the measurement area, TX coordinates
and height, RX height, set of measurements (each tagged by the GPS
coordinates of the location at which it has been acquired).

\emph{Data processing}: the accomplished tasks have been described
in detail in Paragraph \ref{sub:Theory}; consequently, no additional
detail is provided here. 

\emph{Outputs}: RMSE and accuracy evaluated over the test set of SVM
regression and classification, respectively.

\textbf{PM \#2} - \emph{Aim}: for a given position of one or more
data concentrators and in the absence of measurements acquired for
the considered environment, this mode provides the following information
for each concentrator: a) a prediction of its coverage area; b) a
prediction of the RSS at the nodes of rectangular lattice overlapping
the considered environment. In this case, such predictions are based
on classification and regression models generated for environments
\emph{different} from the considered one (and for which measurement
campaigns have been accomplished): for this reason they are called
``blind predictions'' in the following\footnote{The concept of \emph{bootstrapping a propagation model} with measurements
acquired in a given environment and then using it in other environments
is not new in the technical literature (e.g., see \cite[Par. IV-B]{Phillips_2013}
and references therein).}. It should be expected that the predictions generated in this case
are realistic if the adopted classification and regression models
have been originally computed for an environment having a structural
resemblance (in terms of building density and typology, vegetation,
street widths, etc.) with that for which the blind predictions are
evaluated. In practice, this PM can be very useful in radio planning
when assessing the suitability of various options for the positioning
of concentrators in a given area.

\emph{Input data}: geographical data of the considered area, coordinates
and height of the selected data concentrators (transmitters), RX height,
lattice parameters (coordinates of two opposite corners and lattice
step sizes).

\emph{Data processing}: for each available concentrator all the features
are evaluated at the locations corresponding to vertices of the selected
rectangular lattice; then, regression and classification models trained
for environments different from the considered one are used for prediction
over the considered area.

\emph{Outputs}: for each concentrator a RSS prediction over the vertices
of the selected lattice and a prediction of the coverage area are
produced (in particular, RSS prediction results are represented as
a colour map); moreover, a colour map showing, for each point of the
considered area, the best concentrator serving it (i.e., the one from
which the strongest signal is received) is generated (such a map is
called \emph{best server map} and is characterized by as many distinct
colours as the overall number of concentrators, as illustrated in
the following Section).

\textbf{PM \#3} - \emph{Aim}: for a given position of a concentrator,
this mode allows to assess the generalization capability of a given
trained model. In fact, similarly as PM \#2, it generates a RSS blind
prediction, being based on a regression model resulting from data
acquired in environments \emph{different} from (but, hopefully, structurally
similar to) the considered one. However, unlike PM \#2, measurements
are actually available for the considered environment, so that the
accuracy of blind predictions can be assessed at the locations at
which such measurements have been acquired. It is important to point
out that, in this PM, the set of elements forming $S_{test}^{(m)}$
(and consisting of real measurements for the considered area) is further
reduced within the step 2 of the procedure adopted for generating
it (see Paragraph \ref{sub:Theory}); in fact, any datum $\mathbf{x}_{i}\in S_{test}$
such that $d\left(\mathbf{x}_{i}\right)=-1$ (see (\ref{eq:decision_function}))
is discarded. This choice is motivated on the fact that, in this case,
a fair assessment of the prediction accuracy is obtained if only the
data that, on the basis of our classification algorithm, refer to
the locations belonging to the estimated coverage area are taken into
consideration.

\emph{Input data}: geographical data of the measurement area, TX coordinates
and height, RX height, set of measurements (each tagged by the GPS
coordinates of the location in which it has been acquired).

\emph{Data processing}: for the considered concentrator all the features
are evaluated at the positions associated with the considered measurements;
then, regression and classification models trained for environments
different from the considered one are used for RSS prediction over
the considered area.

\emph{Outputs}: coverage area prediction and RSS prediction for the
positions at which the available measurements have been acquired;
RMSE, accuracy, full scale accuracy and percentage of false positives
evaluated on the basis of the available measurements.

\section{Numerical results\label{sec:Numerical-and-experimental}}

In this Section various numerical results generated by our radio planning
tool (called SVMTRP) on the basis of the data sets collected in our
measurement campaigns are illustrated. In such campaigns the following
relevant choices have been made:
\begin{enumerate}
\item The lowest output power level (21 dBm) has been selected for the transmitter
employed in our measurement set-up (see Fig. \ref{fig:Sistema-radio-trasmettitore-rice});
for this reason, the resulting coverage area represents the worst
case associated with a given position of the transmitter.
\item The measurements for the first class of propagation environments (i.e.,
for class \#1) have been acquired in various districts of the cities
of Modena, Bologna and Ferrara, and in relevant areas of the villages
of Carpi and Concordia sulla Secchia (both belonging to the Modena
province). The overall number of measurements collected for class
\#1 is equal to 16009.
\item The measurements for the second class of propagation environments
(i.e., for class \#2) have been acquired in various relevant areas
of the villages of Guiglia, Zocca and Rocca Malatina (all belonging
to the Modena province) and Isera (belonging to the Trento province).
Note that, on the one hand, Guiglia, Zocca and Rocca Malatina are
located in the same upland of the Tuscan-Emilian Apennine, whereas,
on the other hand, Isera is located on the side of a mountain in the
Trentino - Alto Adige region. For this reason, the considered cases
cover the typical land conformation of villages in mountanious/hilly
areas of northern Italy. The overall number of measurements collected
for class \#2 is equal to 11391. 
\item In any propagation environment, when possible, at least one thousand
measurements have been acquired.
\item In the considered cases the 3D maps needed by SVMTRP have been acquired
from the topographic database of the Emilia-Romagna region \cite{Geoportale_ER}
or from that of the autonomous province of Trento \cite{Geoportale_TN}.
All these maps are available in shapefile format.
\end{enumerate}

\subsection{Parameters optimization\label{sub:Param-opt}}

As already mentioned in Section \ref{sub:Theory}, both classification
and regression require the optimization of a couple parameters, namely
$(C,\gamma)$ and $(\tilde{C},\tilde{\gamma})$, respectively. In
our work two different strategies have been adopted to solve this
optimization problem, one for PM \#1, the second for one for PMs \#2
and \#3. This choice is due to the fact that, in the case of PM \#1,
we are interested in adjusting the values of the couples $(C,\gamma)$
and $(\tilde{C},\tilde{\gamma})$ in way that the \emph{classification
and regression accuracies are maximised over test sets}. On the contrary,
in the case of PMs \#2 and \#3, our selection of these parameters
should optimize the \emph{generalization capability} of our SVM-based
methods so that in a given environment reliable prediction can be
evaluated on the basis of a set of measurements acquired in \emph{different}
areas. In practice, in the first case, the \textit{grid-search} method
illustrated in \cite{Hsu} has been adopted. This is based on an exhaustive
search over a finite set of values for the couple $\left(C,\gamma\right)$
(or, equivalently, $(\tilde{C},\tilde{\gamma})$), so that the best
option, in terms of accuracy ($A$) for classification and RMSE for
regression can be found over the available test sets. In the other
case, a grid-search method has been also used, but in a different
way. In fact, first of all, a lower bound on the accuracy (denoted
$A{}_{lb}$ ) and an upper bound on the RMSE (denoted $RMSE_{ub}$)
have been fixed. Then, in the optimisation of the classification (regression)
technique we have searched for, over the given grid, all the values
of the couple $\left(C,\gamma\right)$ ($(\tilde{C},\tilde{\gamma})$)
ensuring an accuracy (RMSE) greater than $A{}_{lb}$ (lower than $RMSE_{ub}$).
Finally, among the selected couples, the one characterized by the
smallest value of $\gamma$ has been chosen. Note that in this procedure
it may occur that, in the set of possible values for the couple $\left(C,\gamma\right)$
($(\tilde{C},\tilde{\gamma})$), no one satisfies the inequality $A\geq A{}_{lb}$
($RMSE\leq RMSE_{ub}$ ); in such a case, the value selected for $A{}_{lb}$
($RMSE_{ub}$) has been reduced (increased) according to the simple
mathematical law given below and the grid-search procedure has been
repeated. In our computer simulations the following specific choices
have been made for the grid-search method employed with PMs \#2 and
\#3:
\begin{itemize}
\item The value $A{}_{th}$=75\% (90\%) has been initially selected for
class \#1 (class \#2) environments and, when needed, was reduced subtracting
$x\mbox{\ensuremath{_{step}}}=5$ dBm ($x\mbox{\ensuremath{_{step}}}=5$
dBm) from it; 
\item The value $RMSE_{ub}=8$ dBm has been selected for both class \#1
and class \#2 environments and, when needed, the last value of $RMSE_{ub}$
was replaced with the larger value $\sqrt{RMSE_{ub}^{2}+4}$;
\item The search interval for the optimal value of $C$ ($\tilde{C}$) has
been restricted to $\left[2^{-8},2^{10}\right]$ ($\left[2^{-3},2^{10}\right]$),
whereas that for the optimal value of $\gamma$ ($\tilde{\gamma}$)
to $\left[2^{-8},2^{6}\right]$ ($\left[2^{-8},2^{3}\right]$).
\end{itemize}
These choices have originated from a trial and error method, whose
starting point have been provided by some practical rules illustrated
in \cite{Hsu} and \cite{LIBSVM}.

\subsection{Performance results\label{sub:Perf-res}}

Our software tool has been run for various areas belonging to class
\#1 and class \#2. We have mainly focused on PM \#3, because of its
practical relevance. However, when operating according to this prediction
mode, our classification and regression algorithms for each of considered
areas have been trained using the measurements collected in all the
other areas of the same class (since a blind prediction was accomplished).
The perfomance improvement that may originate from including in the
training set of PM \#3 the measurements acquired in the considered
considered village/city, but not in the same district, has also been
assessed; this modified version of PM \#3 is denoted PM \#$3'$ in
the following. Various performance results referring to the PMs \#3
and \#$3'$ are shown in Table \ref{tab: flat_results} and Table
\ref{tab:hilly_results} for class \#1 and class \#2, respectively
(in Table \ref{tab:hilly_results} Rocca Malatina A and Rocca Malatina
B refer to the same area in the village of Rocca Malatina, but to
different positions of the transmitter used in our measurement campaigns).
From these results it is easily inferred that:
\begin{itemize}
\item A good accuracy (in particular, a limited RMSE) is achieved by our
classification (regression) algorithm in most of the considered cases.
Note that in the tecnical literature about path loss models for cellular
communications at different frequencies it is commonly stated that,
if measurements are fitted to standard path loss models in specific
environments, a RMSE slightly larger than 9-10 dB can be achieved
over a given link (e.g., see \cite{deOliveira_2006} and \cite{Phillips_2011}).
However, it is also known that the last approach has the following
important drawback: the percentage of links over which each fitted
model makes a reliable prediction is quite low (10-15\%). Our results
have evidenced that our approach does not suffer from this problem
and that, in some cases, the RMSE is even below 9 dB.
\item In various environments, the full-scale accuracy $A_{fs}$ is not
far from the accuracy $A$; this means that the classification capability
does not undergo a significant degradation as we approach the border
of coverage area.
\item The values of the parameter $P_{fp}$ are limited in most of the considered
scenarios.
\item The geographical areas in which the worst results in terms of accuracy
or RMSE are found with PM \#3 (e.g., Concordia exhibits a small accuracy)
are those significantly deviating, in terms of macroscopic characteristics
of the propagation environment, from all the others areas included
in the same class (i.e. class \#1). For this reason, the classification
and regression algorithms are trained by means of a data set that
does not completely match with the considered environments. 
\item In some cases PM \#3 performs slightly better than PM \#$3'$; this
derives from the fact that the exploitation of measurements collected
in the same village/city which the considered environment belongs
to does not necessarily improve accuracy, since distinct districts
of such a village/city may exhibit substantially different propagation
conditions.
\end{itemize}
Finally, further numerical results, not shown here for space limitations,
have evidenced that, in various cases, the performance results obtained
with PM \#1 and PM \#3 for the same area are very close. This leads
to the conclusion that the SVMTRP is an efficient tool for radio planning
and that, in particular, it can generate reliable blind predictions.

\begin{table}
\begin{centering}
\begin{tabular}{|c|c|c|c|c|c|}
\hline 
Radio planning area & PM \# & $A$ & $RMSE$ & $A_{fs}$ & $P_{fp}$\tabularnewline
\hline 
Bologna  & 3 & 82.26  & 7.70  & 82.19  & 17.47 \tabularnewline
\hline 
Concordia  & 3 & 66.22  & 6.77  & 58.62  & 8.13 \tabularnewline
\hline 
Modena & $3'$  & 90.80  & 7.53  & 91.36  & 16.16 \tabularnewline
\hline 
Modena & 3 & 91.03  & 13.44  & 94.20  & 30.11 \tabularnewline
\hline 
\end{tabular}
\par\end{centering}

\caption{Performance results for even areas (class \#1).}

\label{tab: flat_results}
\end{table}

\begin{table}
\begin{centering}
\begin{tabular}{|c|c|c|c|c|c|}
\hline 
Radio planning area & PM \# & $A$ & $RMSE$ & $A_{fs}$ & $P_{fp}$\tabularnewline
\hline 
Rocca Malatina A & $3'$  & 87.31  & 10.27  & 78.79  & 5.62 \tabularnewline
\hline 
Rocca Malatina B & $3'$  & 81.79  & 9.75  & 84.57  & 20.98 \tabularnewline
\hline 
Rocca Malatina A & 3  & 87.94  & 9.93  & 80.96  & 6.73 \tabularnewline
\hline 
Rocca Malatina B & 3  & 82.04  & 10.45  & 89.14  & 23.62 \tabularnewline
\hline 
Isera & 3  & 83.30  & 9.08  & 86.85  & 21.40 \tabularnewline
\hline 
\end{tabular}
\par\end{centering}

\caption{Performance results for hilly/mountanious areas (class \#2).}

\label{tab:hilly_results}
\end{table}

\subsection{Radio coverage estimation\label{sub:radio-coverage}}

\begin{figure}
\begin{centering}
\includegraphics[width=0.85\columnwidth]{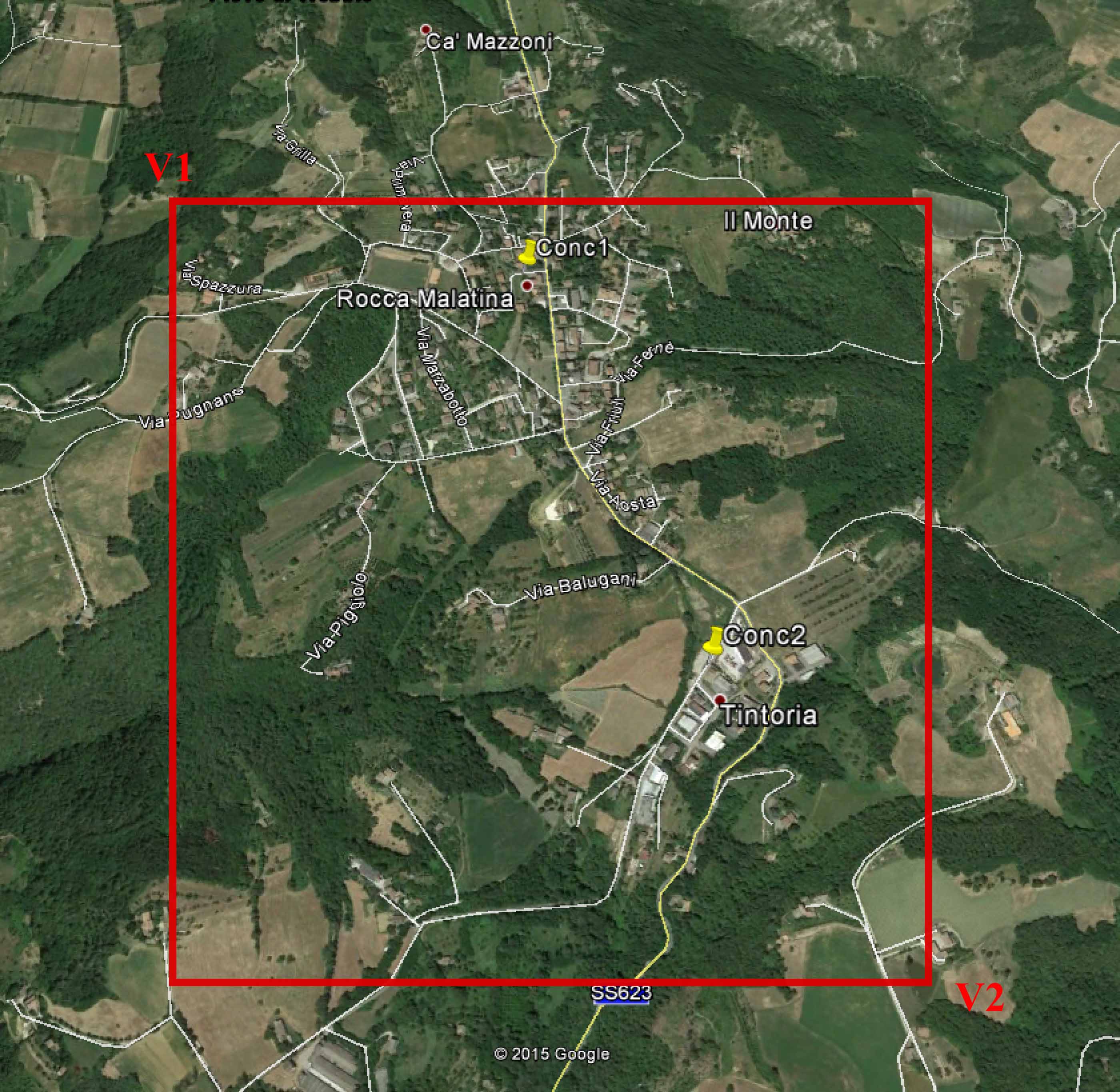}
\par\end{centering}

\caption{Geographic area considered in our blind prediction of radio coverage
in Rocca Malatina.}

\label{fig:example_area}
\end{figure}

\begin{figure}
\begin{centering}
\includegraphics[scale=0.2]{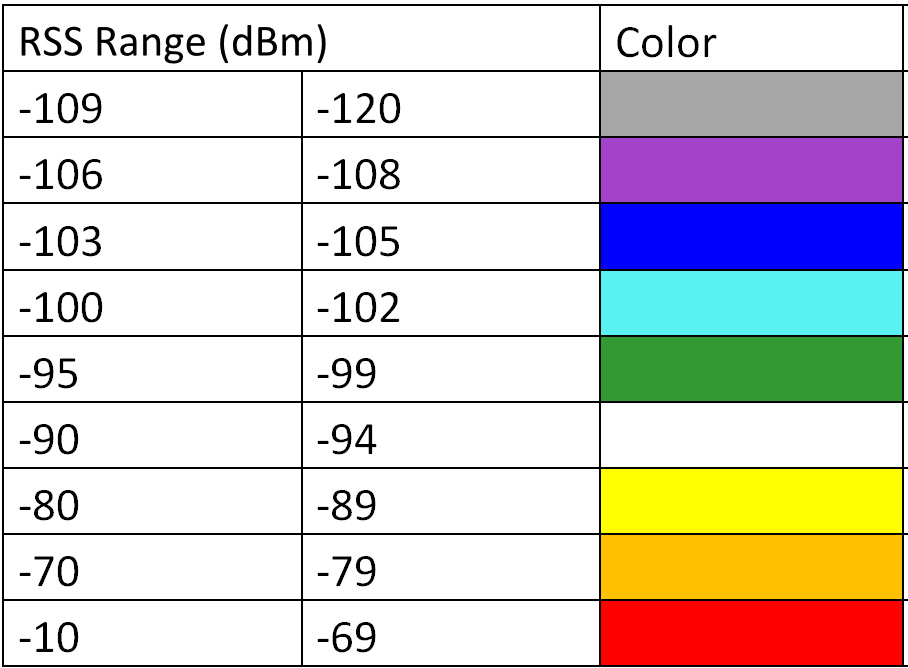}
\par\end{centering}

\caption{Chromatic scale adopted in our signal strength maps.}

\label{fig:colors}
\end{figure}

In developing our SVMTRP, substantial attention has been also paid
to its capability of generating useful graphical outputs for radio
planning when PM \#2 is employed. In the following we show some results
referring to the village of Rocca Malatina, in which two data concentrators
have been placed at specific locations (the coordinates and heights
of these devices are listed in Table \ref{tab:example_coordinates});
the area taken into consideration for radio planning and the geographical
coordinates of the concentrators inside it are shown in Fig. \ref{fig:example_area}.
Note also that the RSS inside the considered area is estimated at
the vertices of a square lattice, whose step size is equal to 8 m.
The resulting outputs are shown in Figs. 7 and \ref{fig:example_results_bestserver}.
In particular, Figs. \ref{fig:results_a} and \ref{fig:results_b}
show the predicted RSS maps for the concentrator \# 1 and concentrator
\# 2, respectively; note that these figures are both based on the
chromatic scale shown in Fig. \ref{fig:colors}. Merging the data
shown in Figs. \ref{fig:results_a} and \ref{fig:results_b} produces
Fig. \ref{fig:results_c}, in which the colour selected for each point
of the delimited area is associated with the strongest signal received
from the two concentrators. The last graphical output is represented
by Fig. \ref{fig:example_results_bestserver}, which shows, for each
point of the delimited area, the best server (i.e., the transmitter
providing the strongest received signal); in particular, the area
covered by the blue (red) color is associated with concentrator \#1
(\#2).

\begin{table}
\begin{centering}
\begin{tabular}{|c|c|c|c|}
\hline 
 & Latitude & Longitude & Height\tabularnewline
\hline 
Concentrator \# 1 & 44.389708\textdegree{} & 10.965048\textdegree{} & 3 m\tabularnewline
\hline 
Concentrator \# 2 & 44.382775\textdegree{} & 10.969014\textdegree{} & 3 m\tabularnewline
\hline 
Vertix \#1 (V1) & 44.391575\textdegree{} & 10.956505\textdegree{} & \multicolumn{1}{c}{}\tabularnewline
\cline{1-3} 
Vertix \#2 (V2) & 44.377973\textdegree{} & 10.973155\textdegree{} & \multicolumn{1}{c}{}\tabularnewline
\cline{1-3} 
\end{tabular}
\par\end{centering}

\caption{Geographic coordinates of the transmitters employed in our measurement
campaigns in the Rocca Malatina area; the coordinates of two opposite
vertices of the rectangular domain considered in our simulations for
radio planning are also given.}

\label{tab:example_coordinates}
\end{table}

\begin{center}
\begin{figure}
\begin{centering}
\subfloat[Signal strength map for concentrator \#1.]{\begin{centering}
\includegraphics[width=0.85\columnwidth]{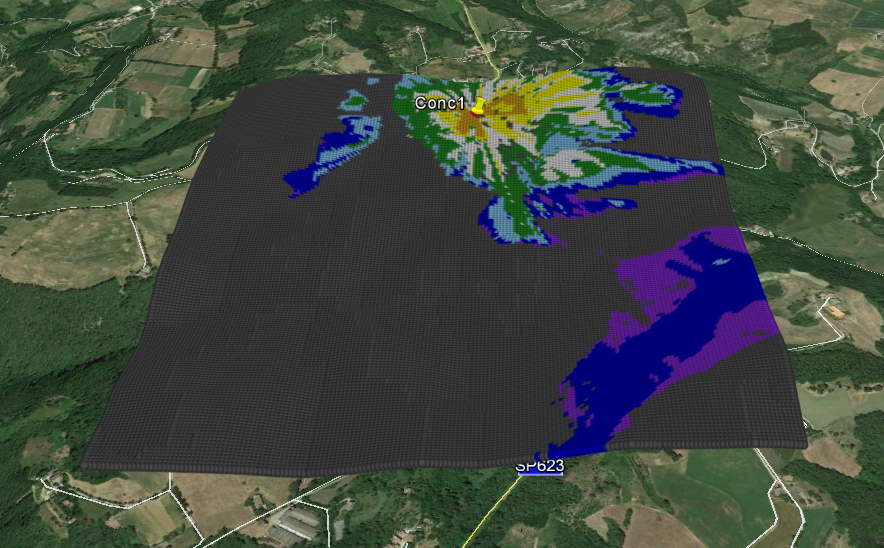}
\par\end{centering}

\begin{centering}

\par\end{centering}

\label{fig:results_a}}
\par\end{centering}

\begin{centering}
\subfloat[Signal strength map for concentrator \#2.]{\begin{centering}
\includegraphics[width=0.85\columnwidth]{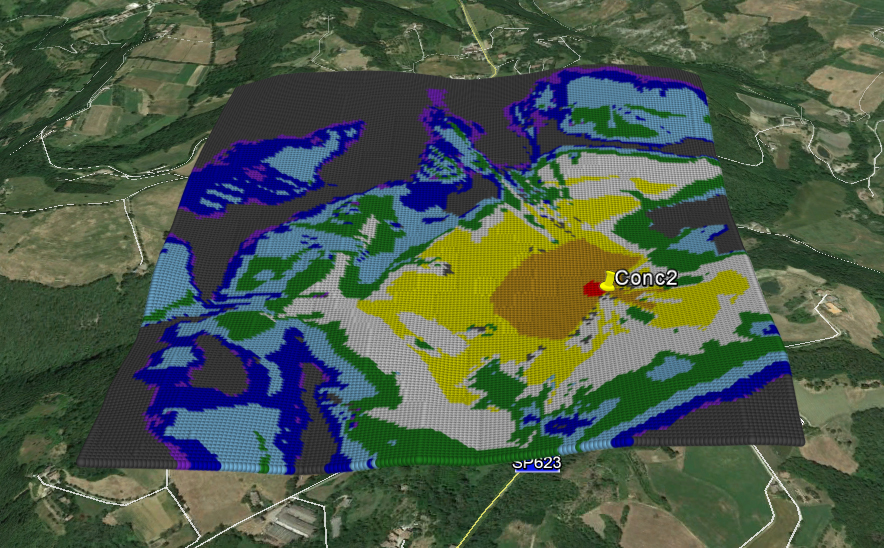}
\par\end{centering}

\begin{centering}

\par\end{centering}

\label{fig:results_b}}
\par\end{centering}

\begin{centering}
\subfloat[Signal strength map for both concentrators. The colour selected for
each point is associated with the strongest signal received from the
two concentrators.]{\centering{}\includegraphics[width=0.85\columnwidth]{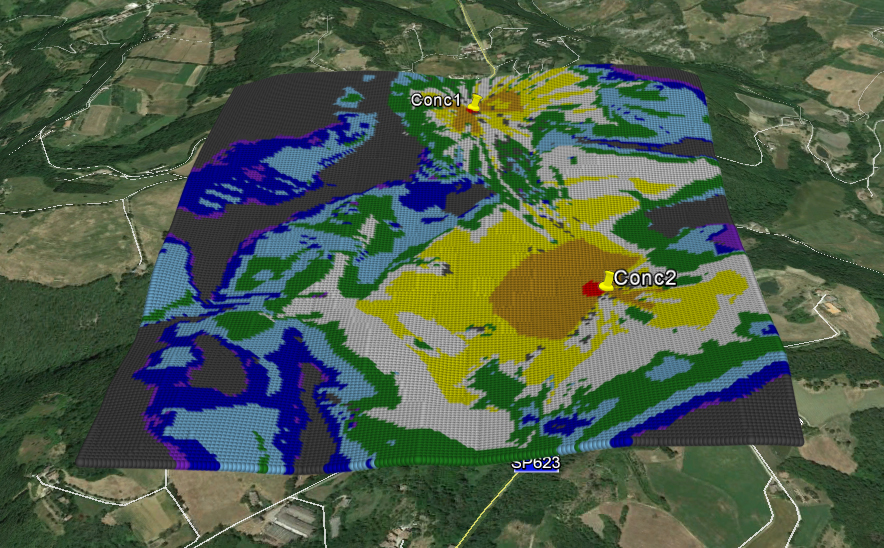}\label{fig:results_c}}\caption{Example of graphical outputs for PM \#2; the Rocca Malatina village
is considered.}

\par\end{centering}

\label{fig:example_result_scale}
\end{figure}

\par\end{center}

\begin{figure}
\begin{centering}
\includegraphics[width=0.85\columnwidth]{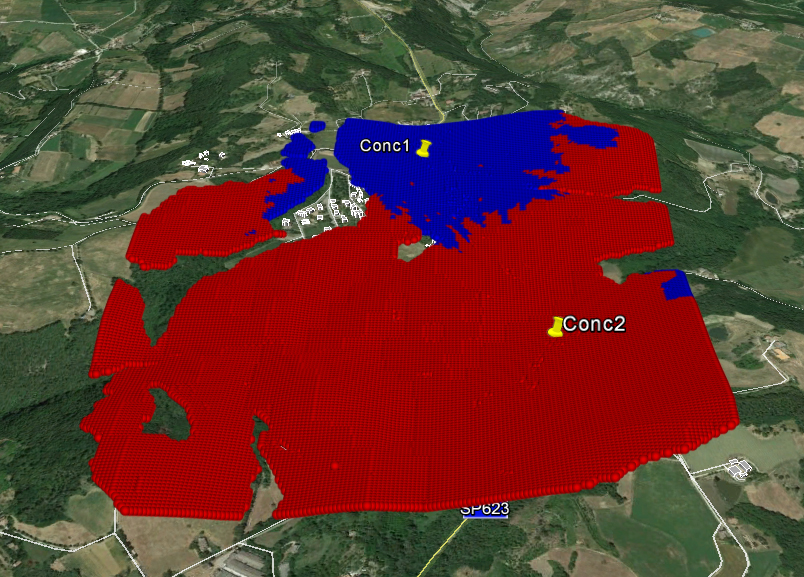}
\par\end{centering}

\caption{Best server output in PM \#2. The Rocca Malatina village is considered;
concentrator \#1 (\#2) is represented by the blue (red) color. The
absence of these colors in some parts of the considered area indicates
that no useful signal is received from the two concentrators.}

\label{fig:example_results_bestserver}
\end{figure}

\section{Conclusions\label{sec:Conclusions}}

In this manuscript SVM-based method for the planning of a radio network
deployed for smart metering and operating in the 169 MHz band has
been developed and its implementation in the MATLAB environment has
been discussed. The efficacy of the devised method has been assessed
in different geographical areas of northern Italy. Our results evidence
that: a) our method achieves a good accuracy at the price of a moderate
computational effort and exploiting a limited set of measurements;
b) it is able to reliably predict the coverage area and the field
strength map in environments for which RSS measurements are unavailable,
but whose structure is similar to that of other environments for which
such measurements have been already acquired. Our future research
work will concern the exploitation of the proposed method for wireless
network planning in other frequency bands.

\section*{Acknowledgment}

We would like to thank Mr. Stefano Bianchi, Mr. Andrea Cavazzoni and
Mr. Tiziano Zocchi (all from CPL Concordia) for their support and
their constructive comments, and Mr. Nicola Tobia (from CPL Concordia
too) for his comments and for his contribution to the development
of the hardware tools employed in our measurements campaigns.

\end{document}